\documentclass[12pt,english,ruled]{article}
\usepackage[T1]{fontenc}
\usepackage[latin9]{inputenc}
\usepackage{geometry}
\usepackage{verbatim}
\usepackage{algorithm2e}
\usepackage{amsmath}
\usepackage{amsthm}
\usepackage{amssymb}
\usepackage{graphicx}
\usepackage{setspace}
\usepackage[authoryear]{natbib}
\usepackage{multirow}

\makeatletter

\providecommand{\tabularnewline}{\\}

\theoremstyle{plain}
\newtheorem{thm}{\protect\theoremname}
\theoremstyle{plain}
\newtheorem{prop}[thm]{\protect\propositionname}

\@ifundefined{date}{}{\date{}}
\usepackage{authblk}

\usepackage{babel}

\usepackage{color}
\usepackage{soul}

\providecommand{\propositionname}{Proposition}
\providecommand{\theoremname}{Theorem}

\usepackage{babel}
\providecommand{\propositionname}{Proposition}
\providecommand{\theoremname}{Theorem}

\@ifundefined{showcaptionsetup}{}{%
 \PassOptionsToPackage{caption=false}{subfig}}
\usepackage{subfig}
\makeatother

\usepackage{babel}
\providecommand{\propositionname}{Proposition}
\providecommand{\theoremname}{Theorem}

\begin{document}
\title{Structured Point Cloud Data Analysis via Regularized Tensor Regression
for Process Modeling and Optimization}
\author{{Hao Yan$^{1}$, Kamran Paynabar$^{2}$, and Massimo Pacella$^{3}$
\\
 $^{1}$School of Computing, Informatics, \& Decision Systems Engineering,
Arizona State University, Tempe, Arizona \\
 $^{2}$School of Industrial and Systems Engineering, Georgia Institute
of Technology, Atlanta, Georgia\\
 $^{3}$Dipartimento di Ingegneria dell'Innovazione, University of
Salento, Lecce, Italy} }
\maketitle
\begin{abstract}
Advanced 3D metrology technologies such as Coordinate Measuring Machine
(CMM) and laser 3D scanners have facilitated the collection of massive
point cloud data, beneficial for process monitoring, control and optimization.
However, due to their high dimensionality and structure complexity,
modeling and analysis of point clouds are still a challenge. In this
paper, we utilize multilinear algebra techniques and propose a set
of tensor regression approaches to model the variational patterns
of point clouds and to link them to process variables. The performance
of the proposed methods is evaluated through simulations and a real
case study of turning process optimization. \vspace{0.2cm}

\end{abstract}

\section{Introduction \label{sec: Introduction}}

Modern measurement technologies provide the means to measure high
density spatial and geometric data in three-dimensional (3D) coordinate
systems, referred to as \textit{point clouds}. Point cloud data analysis
has broad applications in advanced manufacturing and metrology for
measuring dimensional accuracy and shape analysis, in geographic information
systems (GIS) for digital elevation modeling and analysis of terrains,
in computer graphics for shape reconstruction, and in medical imaging
for volumetric measurement to name a few.

\begin{figure}
\centering\includegraphics[width=0.95\linewidth]{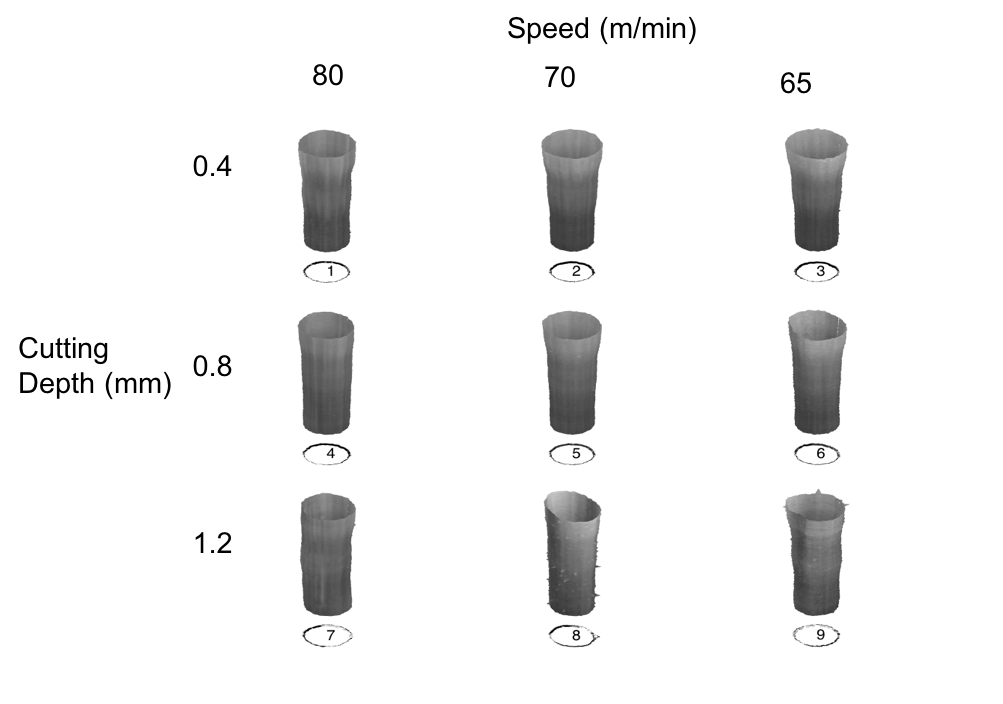}

\caption{Examples of cylindrical surfaces in $9$ different settings. 3D colored
parametric diagram (darker color refers to a minor deviation, brighter
color to a major deviation). Scale 250:1}

\label{Fig: surface} 
\end{figure}

The role of point cloud data in manufacturing is now more important
than ever, particularly in the field of smart and additive manufacturing
processes, where products with complex shape and geometry are manufactured
with the help of advanced technologies \citep{gibson2010additive}.
In these processes, the dimensional and geometric accuracy of manufactured
parts are measured in the form of point clouds using modern sensing
devices, including touch-probe coordinate measuring machines (CMM)
and optical systems, such as laser scanners. Modeling the relationship
of the dimensional accuracy, encapsulated in point clouds, with process
parameters and machine settings is vital for variation reduction and
process optimization. As an example, consider a turning process where
the surface geometry of manufactured parts is affected by two process
variables; namely cutting speed and cutting depth. Figure \ref{Fig: surface}
shows nine point-cloud samples of cylindrical parts produced with
different combinations of the cutting speed and cutting depth. Each
point-cloud sample represents the dimensional deviation of a point
from the corresponding nominal value. The main goal of this paper
is to propose novel tensor regression methods to quantify the relationship
between structured point clouds and scalar predictors. In this paper,
we focus on a class of point clouds where the measurements are taken
on a pre-specified grid. We refer to this as structured point-cloud,
commonly found in dimensional metrology \citep{pieraccini20013d,colosimo2010quality}.
The widely used metrology system to acquire a structured point-cloud
is a conventional CMM, where points can be sampled one by one (acquisition
in a single point mode). In such systems, the localization of the
points on a surface can be accurately controlled by the operator.
Due to the optimal traceability of CMMs, high accuracy and precision
are obtained, making a CMM one of the most important metrology systems
in manufacturing.

Point-cloud data representation and analysis for surface reconstruction
have been extensively discussed in the literature. Point clouds are
often converted to polygon or triangle mesh models \citep{baumgart1975winged},
to NURBS surface models \citep{rogers2000introduction}, or to CAD
models by means of surface reconstruction techniques, such as Delaunay
triangulation \citep{shewchuk1996triangle}, alpha shapes \citep{edelsbrunner1994three},
and ball pivoting \citep{bernardini1999ball}. Although these techniques
are effective in providing a compact representation of point clouds,
they are not capable of modeling the relationship between point clouds
and some independent variables. In the area of process improvement,
the literature on modeling and analysis of point clouds can be classified
into two general categories :\emph{ }\textit{(i) process monitoring
}\textit{\emph{and }}\textit{(ii) process modeling and optimization}\textit{\emph{
approaches.}}

\textit{\emph{Research in the process monitoring category mainly focuses
on finding unusual patterns in the point cloud to detect out-of-control
states and corresponding assignable causes, e.g., combining parametric
regression with univariate and multivariate control charts for quantifying
3D surfaces with spatially correlated noises. However, these models
assume that a parametric model exists for 3D point clouds, which may
not be true for surfaces with complex shapes. To address this challenge,
\citet{wells2013statistical} proposed to use Q--Q plots to transform
the high-dimensional point cloud monitoring into a linear profile
monitoring problem. This approach, however, fails to capture the spatial
information of the data as it reduces a 3D point cloud to a Q-Q plot.
\citet{colosimo2014profile} applied the Gaussian process to model
and monitor 3D surfaces with spatial correlation. However, applying
Gaussian process models can be inefficient for high-dimensional data
such as those in our application. The main objective of the second
category, namely, process modeling and optimization, is to build a
response surface model of a structured point cloud as a function of
some controllable factors. This model is then used to find the optimal
control settings to minimize the dimensional and geometric deviations
of produced parts from nominal values.}} Despite the importance of
the topic, little research can be found in this area. \citet{colosimo2011analyzing}
proposed a two-step approach where first a high-dimensional point
cloud is reduced to a set of features using complex \textit{\emph{Principal
Component Analysis (PCA) and then, the multivariate ANOVA is applied
to the extracted features to test the effect of a set of scalar factors
on the features mean. The main issue of the two-step approach is that
the dimension reduction and modeling/test steps are carried out separately.
Hence, the relationship of scalar factors with point cloud is not
considered in the dimension reduction step. }}

Because of its grid structure, a structured point cloud can be compactly
represented in a multidimensional array also known as a tensor. Therefore,
modeling the structured point cloud as a function of some controllable
factors can be considered as a tensor regression problem. The objective
of this paper is to develop a tensor regression framework to model
a tensor response as a function of some scalar predictors and use
this model for prediction and process optimization. The main challenge
in achieving this objective is the high dimensionality resulting in
a large number of parameters to be estimated.

To achieve this, we take the advantage of the fact that the essential
information of high-dimensional point cloud data lies in a low-dimensional
space and we use low-dimensional basis to significantly reduce the
dimensionality and the number of models parameters. To select appropriate
bases, we introduce two approaches, namely One-step Tensor Decomposition
Regression (OTDR) and Regularized Tensor Regression (RTR). OTDR is
a data-driven approach, where the basis and coefficients are learned
automatically from the tensor regression objective function. In RTR,
we project the response tensor on a set of predefined basis such as
Splines with roughness penalization to control the smoothness of the
estimated tensor response.

The remainder of the paper is organized as follows. Section 2 gives
a brief literature review on functional and tensor regression models.
Section 3 provides an overview of the basic tensor notations and multilinear
algebra operations. Section 4 first introduces the general regression
framework for tensor response data and then elaborates the two approaches
for basis selection, i.e., OTDR and RTR. Section 5 validates the proposed
methodology by using simulated data with two different types of structured
point clouds. In this section, the performance of the proposed method
is compared with existing two-step methods in terms of the estimation
accuracy and computational time. In Section 6, we illustrate a case
study for process modeling and optimization in a turning process.
Finally, we conclude the paper with a short discussion and an outline
of future work in Section 7.

\section{Literature Review}

We review three categories of models in this area. The first category
focuses on regression models with a multivariate vector response.
For example, similar to \citet{colosimo2016multilinear}, one can
use a multilinear extension of PCA \citep{jolliffe2002principal}
to reduce the dimensionality of a multidimensional array of data (a
tensor response) and then build a regression model for the estimated
PC scores. \citet{reiss2010fast} proposed a function-on-scalar regression,
in which the functional response is linked with scalar predictors
via a set of functional coefficients to be estimated in a predefined
functional space. Other methods such as partial least squares \citep{helland1990partial}
or sparse regression \citet{peng2010regularized} are also capable
of regressing a multivariate or functional\textit{\emph{ response
on scalar predictors. Although these methods are effective for modeling
}}multivariate\textit{\emph{ vectors, they are inadequate for analysis
of structured point clouds due to the ultrahigh dimensionality as
well as their complex tensor structures \citep{zhou2013tensor}. The
second category pertains to regression models with a scalar response
and tensor covariates. For example, \citet{zhou2013tensor} proposed
a regression framework in which the dimensionality of tensor covariates
is substantially reduced by applying low-rank tensor decomposition
techniques leading to efficient estimation and prediction. The third
category, directly related to our problem, is the modeling of a tensor
response as a function of scalar predictors. \citep{penny2011statistical}
independently regressed each entry of the response tensor on scalar
predictors and generated a statistical parametric map of the coefficients
across the entire response tensor. This approach often required a
preprocessing smoothing step to denoise the tensor response. For example,
\citet{li2011multiscale} proposed a multi-scale adaptive approach
to denoise the tensor response before fitting the regression model.
However, the major drawback of this approach is that all response
variables are treated independently and hence, important spatial correlation
is often ignored. To overcome this problem, \citet{li2016parsimonious}
built a parsimonious linear tensor regression model by making the
generalized sparsity assumption about the tensor response. Although
the sparsity assumption is valid in neuroimaging applications, it
may not be valid in other point cloud applications such as the one
discussed in this paper. }}

\section{Basic Tensor Notation and Multilinear Algebra}

In this section, we introduce basic notations, definitions, and operators
in multilinear (tensor) algebra that we use in this paper. Throughout
the paper, scalars are denoted by lowercase italic letters, e.g.,
$a$, vectors by lowercase boldface letters, e.g., $\mathbf{a}$,
matrices by uppercase boldface letter, e.g., $\mathbf{A}$, and tensors
by calligraphic letters, e.g., $\mathcal{A}$. For example, an order-$K$
tensor is represented by $\mathcal{A}\in\mathbb{R}^{I_{1}\times\cdots\times I_{K}}$,
where $I_{k}$ represents the mode-$k$ dimension of $\mathcal{A}$.
The mode-$k$ product of a tensor $\mathcal{A}$ by a matrix $\mathbf{V}\in\mathbb{R}^{P_{k}\times I_{k}}$
is defined by $(\mathcal{A}\times_{k}\mathbf{V})(i_{1},\cdots,i_{k-1},j_{k},i_{k+1},\cdots,i_{K})=\sum_{i_{k}}A(i_{1},\cdots,i_{k},\cdots,i_{K})V(j_{k},i_{k})$.
The Frobenius norm of a tensor $\mathcal{A}$ can be defined as $\|\mathcal{A}\|_{F}^{2}=\sum_{i_{1},\cdots,i_{K}}A(i_{1},\cdots,i_{k},\cdots,i_{K})^{2}$.
The n-mode unfold operator maps the tensor $\mathcal{A}$ into matrix
$\mathbf{A}_{(n)}$, where the columns of $\mathbf{A}_{(n)}$ are
the $n$-mode vectors of $\mathcal{A}$.

Tucker decomposition decomposes a tensor into a core tensor multiplied
by a matrix along each mode, i.e., $\mathcal{A}=\mathcal{S}\times_{1}\mathbf{U}^{(1)}\times_{2}\mathbf{U}^{(2)}\cdots\times_{K}\mathbf{U}^{(K)}$,
where $\mathbf{U}^{(k)}$ is an orthogonal $I_{k}\times I_{k}$ matrix
and is a principal component mode-$k$. Tensor product can be represented
equivalently by a Kronecker product, i.e., $\mathrm{vec}(\mathcal{A})=(\mathbf{U}^{(K)}\otimes\cdots\otimes\mathbf{U}^{(1)})\mathrm{vec}(\mathcal{S})$,
where $\mathrm{vec}$ is the vectorized operator defined as $\mathrm{vec}(\mathcal{A})=\mathbf{A}_{(K+1)}$
($\mathbf{A}_{(K+1)}$ refers to the unfolding of $\mathcal{A}$ along
the additional $(K+1)^{th}$ mode, which is an $I_{1}\times I_{2}\times\cdots\times I_{K}$-dimensional
vector). The definition of Kronecker product is as follow: Suppose
$\mathbf{A}\in\mathbb{R}^{m\times n}$ and $\mathbf{B}\in\mathbb{R}^{p\times q}$
are matrices, the Kronecker product of these matrices, denoted by
$\mathbf{A}\otimes\mathbf{B}$, is an $mp\times nq$ block matrix
given by $\mathbf{A}\otimes\mathbf{B}=\left[\begin{array}{ccc}
a_{11}\mathbf{B} & \cdots & a_{1n}\mathbf{B}\\
\vdots & \ddots & \vdots\\
a_{m1}\mathbf{B} & \cdots & a_{mn}\mathbf{B}
\end{array}\right]$.

\section{Tensor Regression Model with Scalar Input \label{sec:Tensor-Regression-Model}}

In this paper, for simplicity and without loss of generality, we present
our methodology with a 2D tensor response. However, this can be easily
extended to higher order tensors by simply adding other dimensions.
Suppose a training sample of size $N$ is available that includes
tensor responses denoted by $\mathbf{Y}_{i}\in\mathbb{R}^{I_{1}\times I_{2}}$,
$i=1,\cdots,N$ along with the corresponding input variables denoted
by $\mathbf{x}_{i}\in\mathbb{R}^{p\times1}$, $i=1,\cdots,N$, where
$p$ is the number of regression coefficients. The tensor regression
aims to link the response $\mathbf{Y}_{i}$ with the input variables
$\mathbf{x}_{i}$ through a tensor coefficient $\mathcal{A}\in\mathbb{R}^{I_{1}\times I_{2}\times p}$
such that

\begin{equation}
\mathbf{Y}_{i}=\mathcal{A}\times_{3}\mathbf{x}_{i}+\delta\mathbf{E}_{i},i=1,\cdots,N,\label{eq: matrixreg}
\end{equation}
where $\mathbf{E}_{i}$ represents the random noises. We combine the
response data $\mathbf{Y}_{i}$ and the residual $\mathbf{E}_{i}$
across the samples into 3D tensors denoted by $\mathcal{Y}\in\mathbb{R}^{I_{1}\times I_{2}\times N}$
and $\mathcal{E}\in\mathbb{R}^{I_{1}\times I_{2}\times N}$, respectively.
Furthermore, we combine all $\mathbf{x}_{i}$'s into a single input
matrix $\mathbf{X}\in\mathbb{R^{\mathrm{\mathit{N\times p}}}}$, where
the first column of $\mathbf{X}$ is $\mathbf{1}\in\mathbb{R^{\mathrm{\mathit{N\times1}}}}$
corresponding to the intercept coefficients. Therefore, \eqref{eq: matrixreg}
can compactly be represented in the tensor format as shown in \eqref{eq: tensorreg}:
\begin{equation}
\mathcal{Y}=\mathcal{A}\times_{3}\mathbf{X}+\delta\mathcal{E},\label{eq: tensorreg}
\end{equation}
where $\mathcal{E}$ is assumed to follow a tensor normal distribution
as $\mathcal{E}\sim N(0,\Sigma_{1},\Sigma_{2},\Sigma_{3})$ \citep{manceur2013maximum},
or equivalently $e=\mathrm{vec}(\mathcal{E})\sim N(0,\Sigma_{3}\otimes\Sigma_{2}\otimes\Sigma_{1})$.
$\Sigma_{1}$ and $\Sigma_{2}$ represent the spatial correlation
of the noise that are assumed to be defined by $\Sigma_{k}|_{i_{1},i_{2}}=\exp(-\theta\|r_{i_{1}}-r_{i_{2}}\|^{2})=\exp(-\theta\|r_{i_{1}}-r_{i_{2}}\|^{2});k=1,2$.
$\Sigma_{3}$ represents the between-sample variation. We further
assume~the samples are independent, and hence, $\Sigma_{3}$ is a
diagonal matrix, defined by $\Sigma_{3}=diag(\sigma_{1}^{2},\sigma_{2}^{2},\cdots,\sigma_{N}^{2})$,
where $diag$ is the diagonal operator that transforms a vector to
a diagonal matrix with the corresponding diagonal elements. The tensor
coefficients can be estimated by minimizing the negative likelihood
function $\mathbf{a}$, which can be solved by $\hat{\mathcal{A}}=\mathcal{Y}\times_{3}(\mathbf{X}^{T}\Sigma_{3}^{-1}\mathbf{X})^{-1}\mathbf{X}^{T}\Sigma_{3}^{-1}$.
The detailed proof is shown in Appendix \ref{sec:Lieklihoodderive}.
However, since the dimensions of $\mathcal{A}$ is too high, solving
$\hat{\mathcal{A}}$ directly can result in the severe overfitting.
In most practical applications, however, $\mathcal{A}$ lies in a
low-dimensional space. Hence, we assume that $\mathcal{A}$ can be
represented in a low-dimensional space expanded by basis $\mathbf{U}^{(k)},k=1,2$,
as shown in \eqref{eq: coefbasis}: 
\begin{equation}
\mathcal{A}=\mathcal{B}\times_{1}\mathbf{U}^{(1)}\times_{2}\mathbf{U}^{(2)}+\mathcal{E}_{\mathcal{A}},\label{eq: coefbasis}
\end{equation}
where $\mathcal{E}_{\mathcal{A}}$ is the residual tensor of projecting
the coefficient $\mathcal{A}$ into a low-dimensional space. $\mathcal{B}\in\mathbb{R}^{P_{1}\times P_{2}\times p}$
is the core tensor (or coefficient tensor) after the projection. If
$\mathbf{U}^{(k)}$ is complete, the residual tensor $\|\mathcal{E}_{\mathcal{A}}\|_{F}=0$.
As $\mathcal{A}$ is low rank, we can use a set of low-dimensional
bases $\mathbf{U}^{(k)}\in\mathbb{R}^{I_{k}\times P_{k}}$ (i.e.,
$P_{k}\ll I_{k}$) to significantly reduce the dimensionality of the
coefficient tensor $\mathcal{A}$ while keeping $\|\mathcal{E}_{\mathcal{A}}\|_{F}$
close to zero. Since $\mathcal{E}_{\mathcal{A}}$ is negligible, given
$\mathbf{U}^{(1)}$ and $\mathbf{U}^{(2)}$, $\mathcal{A}$ can be
approximated by $\mathcal{B}\times_{1}\mathbf{U}^{(1)}\times_{2}\mathbf{U}^{(2)}$.
Therefore, by combining \eqref{eq: coefbasis} and \eqref{eq: tensorreg},
we have 
\begin{equation}
\mathcal{Y}=\mathcal{B}\times_{1}\mathbf{U}^{(1)}\times_{2}\mathbf{U}^{(2)}\times_{3}\mathbf{X}+\delta\mathcal{E}.\label{eq: onestepTensorreg}
\end{equation}
To estimate $\mathcal{B}$, the following likelihood function can
be derived, as shown in \eqref{eq: onestepreg}. 
\begin{equation}
\min_{\mathcal{B}}l(\mathcal{B})=(\mathbf{y}-(\mathbf{X}\otimes\mathbf{U}^{(2)}\otimes\mathbf{U}^{(1)})\boldsymbol{\beta})^{T}(\Sigma_{3}\otimes\Sigma_{2}\otimes\Sigma_{1})^{-1}(\mathbf{y}-(\mathbf{X}\otimes\mathbf{U}^{(2)}\otimes\mathbf{U}^{(1)})\boldsymbol{\beta})\label{eq: onestepreg}
\end{equation}

\begin{prop}
\label{prop: DataReg} Problem \eqref{eq: onestepreg} has a closed-form
solution expressed by 
\begin{equation}
\hat{\mathcal{B}}=\mathcal{Y}\times_{1}(\mathbf{U}^{(1)T}\Sigma_{1}^{-1}\mathbf{U}^{(1)})^{-1}\mathbf{U}^{(1)T}\Sigma_{1}^{-1}\times_{2}(\mathbf{U}^{(2)T}\Sigma_{2}^{-1}\mathbf{U}^{(2)})^{-1}\mathbf{U}^{(2)T}\Sigma_{2}^{-1}\times_{3}(\mathbf{X}^{T}\Sigma_{3}^{-1}\mathbf{X})^{-1}\mathbf{X}^{T}\Sigma_{3}^{-1}.\label{eq: coefReg}
\end{equation}
\end{prop}

The proof of proposition \ref{prop: DataReg} is shown in Appendix
\ref{sec: scorereg}.

The choice of basis $\mathbf{U}^{(k)},k=1,2$ is important to the
model accuracy. In the next subsections, we propose two methods for
defining basis matrices: The first method Regularized Tensor Regression
(RTR) incorporates the user knowledge about the process and response;
and the second method One-step Tensor Decomposition Regression (OTDR)
is a one-step approach that automatically learns the basis $\mathbf{U}^{(k)},k=1,2$
and coefficients from data.

\subsection{Regularized Tensor Regression (RTR) with Customized Basis Selection}

In some cases, we prefer to customize the basis based on the domain
knowledge and/or data characteristics. For example, a predefined spline
or kernel basis can be used to represent general smooth tensors. Fourier
or periodic B-spline basis can be used to represent smooth tensors
with periodic boundary constraints. Furthermore, a penalty term can
be added to control the level of smoothness. Consequently, the one-step
regression model can be rewritten by 
\begin{equation}
\hat{\boldsymbol{\beta}}=\mathop{\mathrm{argmin}}_{\boldsymbol{\beta}}(\mathbf{y}-(\mathbf{X}\otimes\mathbf{U}^{(2)}\otimes\mathbf{U}^{(1)})\boldsymbol{\beta})^{T}(\Sigma_{3}\otimes\Sigma_{2}\otimes\Sigma_{1})^{-1}(\mathbf{y}-(\mathbf{X}\otimes\mathbf{U}^{(2)}\otimes\mathbf{U}^{(1)})\boldsymbol{\beta})+P(\boldsymbol{\beta}).\label{eq: onestepregpenal}
\end{equation}
However, the dimensions of $\mathbf{X}\otimes\mathbf{U}^{(2)}\otimes\mathbf{U}^{(1)}$
are $NI_{1}I_{2}\times pP_{1}P_{2}$, which is often too large to
compute or even to be stored. To address this computational challenge,
following \citep{yan2015anomaly}, we use a special form of the penalty
term defined by

\begin{equation}
P(\boldsymbol{\beta})=\boldsymbol{\beta}^{T}(\mathbf{X}^{T}\Sigma_{3}^{-1}\mathbf{X})\otimes(\lambda\mathbf{P}_{2}\otimes\mathbf{U}^{(1)^{T}}\Sigma_{1}^{-1}\mathbf{U}^{(1)}+\lambda\mathbf{U}^{(2)^{T}}\Sigma_{2}^{-1}\mathbf{U}^{(2)}\otimes\mathbf{P}_{1}+\lambda^{2}\mathbf{P}_{2}\otimes\mathbf{P}_{1})\boldsymbol{\beta},\label{eq: penalize}
\end{equation}
where $\boldsymbol{\beta}=\mathrm{vec}(\mathcal{B})$, $\mathbf{P}_{k}=(\mathbf{D}_{k}^{2})^{T}\mathbf{D}_{k}^{2}$
is the penalization matrix to control the smoothness among the mode-$k$
of the original tensor. As shown in \citep{xiao2013fast} and \citep{yan2015anomaly},
the penalty term defined with tensor structure works well in the simulation
and achieve the optimal rate of convergence asymptotically under some
mild conditions. In Proposition \ref{prop: twostepReg} We prove that
by using this $P(\mathcal{B})$, not only does Problem \eqref{eq: onestepregpenal}
have a closed-form solution, but also the solution can be computed
along each mode of the original tensor separately, which significantly
reduces the computational complexity. 
\begin{prop}
\label{prop: twostepReg}The optimization problem \eqref{eq: onestepregpenal}
with $P(\mathcal{B})$ defined in \eqref{eq: penalize} can efficiently
be solved via a tensor product given by 
\begin{equation}
\hat{\mathcal{B}}=\mathcal{Y}\times_{1}(\mathbf{U}^{(1)^{T}}\Sigma_{1}^{-1}\mathbf{U}^{(1)}+\lambda\mathbf{P}_{1})^{-1}\mathbf{U}^{(1)^{T}}\Sigma_{1}^{-1}\times_{2}(\mathbf{U}^{(2)^{T}}\Sigma_{2}^{-1}\mathbf{U}^{(2)}+\lambda\mathbf{P}_{2})^{-1}\mathbf{U}^{(2)^{T}}\Sigma_{2}^{-1}\times_{3}(\mathbf{X}^{T}\Sigma_{3}^{-1}\mathbf{X})^{-1}\mathbf{X}^{T}\Sigma_{3}^{-1}.\label{eq: RegTensorRegsol}
\end{equation}
\end{prop}

The proof is given in Appendix \ref{sec:RegTenReg}. The tuning parameter
selection will be discussed in section \ref{subsec: TRS}.

Assuming $I_{1}=I_{2}=I_{0}$, the computational complexity of the
RTR method in each iteration is $O(I_{0}^{2}N^{2}p)$ provided that
the covariance matrix is computed beforehand to save the computational
time in each iteration.

\subsection{One-step Tensor Decomposition Regression (OTDR)}

In cases where little engineering or domain knowledge is available,
it is necessary to learn the basis from the data. Similar to principal
component regression, one technique is to learn the basis from Tucker
decomposition and apply the regression on the Tucker decomposition
core tensor. The details of this method is given in Appendix \ref{sec:Tucker-Decomposition-Regression}.
However, the major limitation of this two-step approach is that the
learned basis may not correspond to the input variables $\mathbf{X}$.

Consequently, in this section, we propose a one-step approach to learn
the basis and coefficients at the same time. To achieve this, we propose
to simultaneously optimize both the coefficient $\boldsymbol{\beta}$
and basis $\mathbf{U}^{(k)}$ in \ref{eq: onestepreg}. Moreover,
instead of the orthogonality constraint on $\mathbf{U}^{(i)},i=1,2$
, we apply the weighted constraint given by $\mathbf{U}^{(k)T}\Sigma_{i}^{-1}\mathbf{U}^{(k)}=I$.
This constraint not only results in a closed-form solution in each
iteration, but also ensures that the estimated basis and residuals
share a similar spatial covariance structure. 
\begin{align}
 & \mathop{\mathrm{argmin}}_{\boldsymbol{\beta},\mathbf{U}^{(k)}}(\mathbf{y}-(\mathbf{X}\otimes\mathbf{U}^{(2)}\otimes\mathbf{U}^{(1)})\boldsymbol{\beta})^{T}(\Sigma_{3}\otimes\Sigma_{2}\otimes\Sigma_{1})^{-1}(\mathbf{y}-(\mathbf{X}\otimes\mathbf{U}^{(2)}\otimes\mathbf{U}^{(1)})\boldsymbol{\beta})\nonumber \\
 & s.t.\quad\mathbf{U}^{(k)T}\Sigma_{k}^{-1}\mathbf{U}^{(k)}=I\label{eq: optimize}
\end{align}
To efficiently optimize \eqref{eq: optimize}, we use the alternative
least square approach, which optimizes $\mathcal{B}$, $\mathbf{U}^{(1)}$,
$\mathbf{U}^{(2)}$, iteratively. From proposition \eqref{prop: DataReg},
we know that if $\mathbf{U}^{(k)},k=1,2$ are given, $\mathcal{B}$
has a closed-form solution as the one in \eqref{eq: onestepreg}.
To optimize $\mathbf{U}^{(k)},k=1,2$, we use the following proposition. 
\begin{prop}
\label{prop: onestepU} Minimizing the negative log-likelihood function
in \eqref{eq: onestepreg} is equivalent to maximizing the projected
scores norm in \eqref{eq: projscore},

\begin{align}
 & \arg\max_{\mathbf{U^{(k)}}}\|\mathcal{Y}\times_{1}\mathbf{U}^{(1)^{T}}\Sigma_{1}^{-1}\times_{2}\mathbf{U}^{(2)^{T}}\Sigma_{2}^{-1}\times_{3}\mathbf{X}_{3}\|^{2};k=1,2,\nonumber \\
 & s.t.\quad\mathbf{U}^{(k)T}\Sigma_{k}^{-1}\mathbf{U}^{(k)}=I,\label{eq: projscore}
\end{align}
where $\mathbf{X}_{3}$ can be computed by the Cholesky decomposition
as $\mathbf{X}_{3}\mathbf{X}_{3}^{T}=\Sigma_{3}^{-1}\mathbf{X}(\mathbf{X}^{T}\Sigma_{3}^{-1}\mathbf{X})^{-1}\mathbf{X}^{T}\Sigma_{3}^{-1}$.
Furthermore, given $\mathbf{U}^{(i)}$, the maximizer of \eqref{eq: projscore}
is given by $\mathbf{U}^{(k)}=\Sigma_{k}^{1/2}\mathbf{\tilde{U}}^{(k)},$
where $\mathbf{\tilde{U}}^{(k)}$ is the first $P_{k}$ eigenvectors
of $\Sigma_{k}^{-1/2}\mathbf{W}_{k}$. $\mathbf{W}_{k}$ is the $k^{th}$
mode unfolding of $\mathcal{W}_{k}=\mathcal{Y}\times_{3-k}\mathbf{U}^{(i)}\Sigma_{3-k}^{-1}\times_{3}\mathbf{X}_{3}$. 
\end{prop}

The procedure for performing OTDR is given in Algorithm 1. The fact
that the resulting sub-problem in each iteration reduces to a generalized
eigenvalue problem significantly speeds up the algorithm. Thus, assuming
that $P_{1}=P_{2}=P_{0}$ and $I_{1}=I_{2}=I_{0}$, the complexity
of the algorithm in each iteration is $O(\min(P_{0}^{2}NI_{0},N^{2}I_{0}^{2}P_{0}))$,
provided that $\Sigma_{k}^{-1},\Sigma_{k}^{1/2},\Sigma_{k}^{-1/2}$
is computed beforehand.

Similar to many non-convex models such as Tucker decomposition and
matrix factorization \citep{kolda2009tensor}, the solution of OTDR
is not unique. For example, it is possible to define an orthogonal
transformation $\mathbf{R}_{k}$ on the basis $\mathbf{U}^{(k)}$
as $\mathbf{U}_{1}^{(k)}=\mathbf{U}^{(k)}\mathbf{R}_{k}$, and $\beta_{1}=(I\otimes\mathbf{R}_{2}^{-1}\otimes\mathbf{R}_{1}^{-1})\beta$
such that the fitted response $\hat{\mathbf{y}}=(\mathbf{X}\otimes\mathbf{U}^{(2)}\otimes\mathbf{U}^{(1)})\boldsymbol{\beta}$
stays the same and the constraint still holds. However, different
matrix factorization and tensor decomposition models are still widely
used despite the non-uniqueness of the solution \citep{kolda2009tensor}.
The reason is that over-complete representation is more flexible and
robust to noise and the fitted response will not be affected \citep{Anandkumar2013}.
However, if uniqueness of parameter is important, it is possible to
add the sparsity constraint on the core tensor so that the algorithm
will search for the rotation to make as many elements $0$ as possible
\citep{Anandkumar2013}. For more detailed discussions about the over-complete
representation in the application of tensor decomposition, please
refer to \citep{Anandkumar2013}.

\begin{algorithm}
\caption{ALS algorithm for RTD}

\begin{itemize}
\item Initialize

Compute $\Sigma_{k}^{-1},\Sigma_{k}^{1/2},\Sigma_{k}^{-1/2}$ and
compute $X_{3}$ through the Cholesky decomposition as $\mathbf{X}_{3}\mathbf{X}_{3}^{T}=\Sigma_{3}^{-1}\mathbf{X}(\mathbf{X}^{T}\Sigma_{3}^{-1}\mathbf{X})^{-1}\mathbf{X}^{T}\Sigma_{3}^{-1}$ 
\item For $i=1,2,\cdots$, $k=1,2$

Compute $\mathcal{W}_{k}=\mathcal{Y}\times_{3-k}\mathbf{U}^{(3-k)^{T}}\Sigma_{3-k}^{-1}\times_{3}X_{3}$,

Compute $\mathbf{W}_{k}$ as the $k$-th mode unfolding of $\mathcal{W}_{k}$.

Compute $\mathbf{\tilde{U}}^{(k)}$ is the first $P_{k}$ eigenvectors
of $\Sigma_{k}^{-1/2}\mathbf{W}_{k}$,

Update$\mathbf{U}^{(k)}=\Sigma_{k}^{1/2}\mathbf{\tilde{U}}^{(k)}$
until convergence 
\item Compute $\mathcal{B}$ based on \eqref{eq: coefReg} 
\end{itemize}
\label{alg: ALSostr} 
\end{algorithm}

\subsection{Tuning Parameter Selection }

\label{subsec: TRS}

In this section, we propose a procedure for selecting the tuning parameters
including the smoothness parameter $\lambda$ in RTD and RTR, and
the covariance parameter $\theta$ in OTDR. To select tuning parameter
$\lambda$ in RTR, since the formulation follows the standard ridge
regression format, the Generalized Cross Validation (GCV) criterion
can be used. That is, $\hat{\lambda}=\arg\min_{\lambda}\mathrm{GCV}(\lambda)=\arg\min_{\lambda}\frac{\|\mathcal{Y}-\hat{\mathcal{B}}\times_{1}\mathbf{U}^{(1)}\times_{2}\mathbf{U}^{(2)}\times_{3}\mathbf{X}\|^{2}/N}{(1-N^{-1}\mathrm{tr}(\hat{\mathbf{H}}_{1}(\lambda))\mathrm{tr}(\hat{\mathbf{H}}_{2}(\lambda))\mathrm{tr}(\hat{\mathbf{H}}_{3}(\lambda)))^{2}}$,
where $\hat{\mathbf{H}}_{k}(\lambda)=\mathbf{U}^{(k)}(\mathbf{U}^{(k)^{T}}\mathbf{U}^{(k)}+\lambda\mathbf{P}_{k})^{-1}\mathbf{U}^{(k)^{T}}$,
$k=1,2$, and $\hat{\mathbf{H}}_{3}(\lambda)=\mathbf{X}(\mathbf{X}^{T}\mathbf{X})^{-1}\mathbf{X}^{T}$.

In the OTDR model, we propose to find the set of tuning parameters
including the covariance parameters $\theta$, $\sigma$, and number
of PCs $P_{1}$, $P_{2}$ by minimizing the BIC criterion defined
in \eqref{eq: covopt-1}, where $P_{1}$ and $P_{2}$ are number of
bases in $\mathbf{U}^{(1)}$ and $\mathbf{U}^{(2)}$. 
\begin{equation}
\min_{\theta,\sigma,P_{1},P_{2}}BIC=N\ln(\frac{1}{N}(\mathbf{y}-(\mathbf{X}\otimes\mathbf{U}^{(2)}\otimes\mathbf{U}^{(1)})\boldsymbol{\beta})^{T}(\Sigma_{3}\otimes\Sigma_{2}\otimes\Sigma_{1})^{-1}(\mathbf{y}-(\mathbf{X}\otimes\mathbf{U}^{(2)}\otimes\mathbf{U}^{(1)})\boldsymbol{\beta})+(P_{1}+P_{2})p\ln N\label{eq: covopt-1}
\end{equation}

\section{Simulation Study}

In this section, we conduct simulations to evaluate the performance
of the proposed OTDR and RTR for structured point cloud modeling.
We simulate $N$ structured point clouds as training samples $\mathbf{Y}_{i},i=1,\cdots,N$
for two different responses, namely, a wave-shape response surface
and a truncated cone response. The simulated data is generated according
to $\mathbf{Y}_{i}=\mathbf{M}+\mathbf{V}_{i}+\mathbf{E}_{i}$, or
equivalently in the tensor form, $\mathcal{Y}=\mathcal{M}+\mathcal{V}+\mathcal{E}$,
where $\mathcal{Y}$ is a 3rd order tensor combining $\mathbf{Y}_{i},i=1,\cdots,N$,
$\mathcal{M}$ is the mean of the point cloud data and $\mathcal{V}$
is the variational pattern of the point cloud due to different levels
of input variables, $\mathbf{x}_{i}$. $\mathcal{E}$ is a tensor
of random noises. We consider two cases to generate noises: 1) i.i.d
noise, where $\mathbf{E}_{i}\overset{i.i.d}{\sim}N(0,\sigma^{2})$;
and 2) non-i.i.d noise, $e=\mathrm{vec}(\mathcal{E})\sim N(0,\Sigma_{3}\otimes\Sigma_{2}\otimes\Sigma_{1})$.
Since the point cloud is spatially correlated, we put the spatial
correlation structure on the covariance matrix on two spatial axes
$\Sigma_{1},\Sigma_{2}$, i.e., $\Sigma_{1}|_{i_{1},i_{2}}=\Sigma_{2}|_{i_{1},i_{2}}=\exp(-\theta\|r_{i_{1}}-r_{i_{2}}\|^{2})$.

\paragraph{Case 1. Wave-shape surface point cloud simulation }

\begin{figure}
\subfloat[Case 1 with i.i.d noise]{\centering\includegraphics[width=0.46\linewidth]{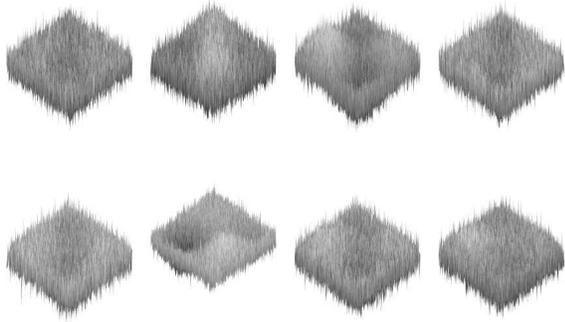}

}\hfill{}\subfloat[Case 2 with i.i.d noise]{\centering\includegraphics[width=0.46\linewidth]{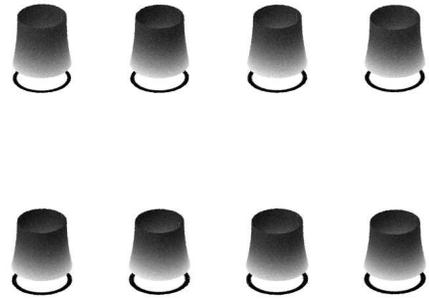}

}

\subfloat[Case 1 with non-i.i.d noise]{\centering\includegraphics[width=0.46\linewidth]{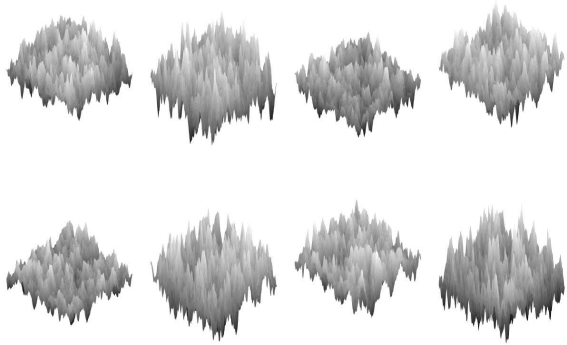}

}\hfill{}\subfloat[Case 2 with non-i.i.d noise]{\centering\includegraphics[width=0.46\linewidth]{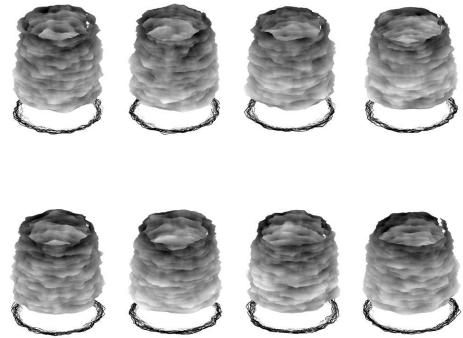}

}

\caption{Examples of generated point cloud for simulation study with i.i.d
and non i.i.d noise}

\label{Fig: simexample} 
\end{figure}

We simulate the surface point cloud in a 3D Cartesian coordinate system
$(x,y,z)$ where $0\leq x,y\leq1$. The corresponding $z_{i_{1}i_{2}}$
values at $(\frac{i_{1}}{I_{1}},\frac{i_{2}}{I_{2}}),i_{1}=1,\cdots,I_{1};i_{2}=1,\cdots I_{2}$,
with $I_{1}=I_{2}=200$ for $i^{th}$ sample recorded in the matrix
$\mathbf{Y}_{i}$, is generated by $\mathcal{Y}=\mathcal{V}+\mathcal{E}$.
We simulate the variational patterns of point cloud surface $\mathcal{V}$,
according to the following linear model $\mathcal{V}=\mathcal{B}\times_{1}\mathbf{U}^{(1)}\times_{2}\mathbf{U}^{(2)}\times_{3}\mathbf{X}$.
In the simulation setup, we select three basis matrices, namely $\mathbf{U}^{(k)}=[\mathbf{u}_{1}^{(k)},\mathbf{u}_{2}^{(k)},\mathbf{u}_{3}^{(k)}]$
with $\mathbf{u}_{\alpha}^{(k)}=[\sin(\frac{\pi\alpha}{n}),\sin(\frac{2\pi\alpha}{n}),\cdots,\sin(\frac{n\pi\alpha}{n})]^{T},\alpha=1,2,3$.
The two mode-3 slices of $\mathcal{B}\in\mathbb{R}^{3\times3\times2}$
is generated as $\mathbf{B}_{1}=\left[\begin{array}{ccc}
4 & 1 & 0\\
1 & 0.1 & 0\\
1 & 0 & 1
\end{array}\right]$, $\mathbf{B}_{2}=\left[\begin{array}{ccc}
1 & 2 & 0\\
1 & 3 & 0\\
1 & 0 & 0.2
\end{array}\right]$. The input matrix $\mathbf{X}$ are randomly sampled from the standard
normal distribution $N(0,1)$. In this study, we generate 100 samples
according the foregoing procedure. The examples of the generated point
cloud surface with the i.i.d noise and non-i.i.d noise ($\theta=10$)
are shown in Figure \ref{Fig: simexample}.

\paragraph{Case 2. Truncated cone point cloud simulation}

We simulate truncated cone point clouds in a 3D cylindrical coordinate
system $(r,\phi,z)$, where $\phi\in[0,2\pi]$, $z\in[0,1]$. The
corresponding $r$ values at $(\phi,z)=(\frac{2\pi i_{1}}{I_{1}},\frac{i_{2}}{I_{2}})$,
$i_{1}=1,\cdots,I_{1};i_{2}=1,\cdots I_{2}$ with $I_{1}=I_{2}=200$
for the $i^{th}$ sample are recorded in the matrix $\mathbf{Y}_{i}$.
We simulate the variational patterns of point cloud surface $\mathcal{V}$
according to $r(\phi,z)=\frac{r+z\tan\theta}{\sqrt{1-e^{2}\cos^{2}\phi}}+c(z^{2}-z)$
with three different settings ($0.9\times$, $1\times$, and $1.1\times$)
of the normal setting, i.e., $\theta_{0}=\frac{\pi}{8},r_{0}=1.3,e_{0}=0.3,c_{0}=0.5$
, which corresponds to 1) different angles of the cone; 2) different
radii of the upper circle; 3) different eccentricities of top and
bottom surfaces; and 4) different curvatures of the side of the truncated
cone. Furthermore, we define four input variables by $x_{1}=\tan\theta$,
$x_{2}=r$, $x_{3}=e^{2}$, $x_{4}=c$ and record them in an input
matrix $\mathbf{X}$ of size $81\times4$. These nonlinear transformations
lead to a better linear approximation of the point cloud in the cylindrical
coordinate system given the input matrix $\mathbf{X}$. Finally, we
use a full factorial design to generate $3^{4}=81$ training samples
with different combinations of the coefficients. The examples of the
generated truncated cones with i.i.d noise and non-i.i.d noise ($\theta=10$)
are shown in Figure \ref{Fig: simexample}. We simulated 1000 test
examples $\mathcal{Y}_{te}$ based on different settings of $\theta,r,e,c$
(uniformly between the lowest and highest settings in the design table).

For both cases, the goal is to find the relationship between the point
cloud tensor $\mathcal{Y}$ and input variables $\mathbf{X}$. We
compare our proposed RTR and OTDR with two existing methods in the
literature. The benchmark methods we used for comparison include vectorized
principal component analysis regression (VPCR), Tucker decomposition
regression (TDR) and simple linear regression (LR). For VPCR, PCA
is applied on the unfolded matrix denoted by $\mathbf{Y}_{(3)}$.
In LR, we separately conduct a linear regression for each entry of
the tensor $\mathcal{Y}$ with the input variables $\mathbf{X}$.
In TDR, we use basis matrices learned from Tucker Decomposition of
the data. For RTR, we use B-spline with 20 knots on each dimension.
It should be noted that in Case 2, we apply the periodic B-spline
with the period of $2\pi$ to model the periodicity in the $\theta$
direction. In OTDR, the basis is automatically learned from the data.
The tuning parameters of RTR and OTDR are selected by using the GCV
and BIC criteria. For Case 1 and 2, the relative sum of squared error
(SSE) between $\mathcal{Y}_{te}$ and the predicted tensor $\hat{\mathcal{Y}}_{te}$
defined by $\frac{\|\mathcal{Y}_{te}-\hat{\mathcal{Y}}_{te}\|^{2}}{\|\hat{\mathcal{Y}}_{te}-\bar{\mathcal{Y}}_{te}\|^{2}}$
is computed from 10,000 simulation replications under different noise
levels, $\delta$, and two noise structures: i.i.d and non-i.i.d with
different $\theta$ values. The results are reported in Tables \ref{Table: SSE1noniid}.
Furthermore, the average computational time per sample for each method
is reported: RTR, 1.19s; OTDR, 1.7s; VPCR, 0.97s; and LR, 0.77s. It
can be seen that the proposed RTR and OTDR have a similar level of
complexity to VPCR. Recall that the complexities of RTR, OTDR, and
VPCR are $O(I_{0}^{2}N^{2}p)$, $O(\min(P_{0}^{2}NI_{0},N^{2}I_{0}^{2}P_{0}))$,
and $O(I_{0}^{2}N^{2})$ respectively. As $p$ and $P_{0}$ are often
small, the complexity of these methods are roughly the same.

\begin{table}
\caption{SSEs (Unit: percentage \%) of the proposed methods when $\sigma=1$
with i.i.d and non-i.i.d noise ($\theta=10$)}

\centering

{\small{}}%
\begin{tabular}{c|cc|cc|cc|cc}
\hline 
 & \multicolumn{4}{c|}{{\small{}Case 1}} & \multicolumn{4}{c}{{\small{}Case 2}}\tabularnewline
\hline 
 & \multicolumn{2}{c|}{{\small{}Non i.i.d}} & \multicolumn{2}{c|}{{\small{}i.i.d}} & \multicolumn{2}{c|}{{\small{}Non i.i.d}} & \multicolumn{2}{c}{{\small{}i.i.d}}\tabularnewline
\hline 
{\small{}$\delta=0.1$ } & {\small{}Mean } & {\small{}STD } & {\small{}Mean } & {\small{}STD } & {\small{}Mean } & {\small{}STD } & {\small{}Mean } & {\small{}STD}\tabularnewline
\hline 
{\small{}RTR } & {\small{}136.79 } & {\small{}6.69 } & {\small{}0.36 } & {\small{}0.00048 } & {\small{}2.91 } & {\small{}0.24 } & {\small{}0.0012 } & {\small{}4.e-06 }\tabularnewline
{\small{}TDR } & {\small{}57.44 } & {\small{}29.96 } & {\small{}0.00043 } & {\small{}1e-05 } & {\small{}3.08 } & {\small{}0.71 } & {\small{}9e-05 } & {\small{}4e-07}\tabularnewline
{\small{}VPCR } & {\small{}60.72 } & {\small{}27.39 } & {\small{}0.028 } & {\small{}9e-05 } & {\small{}2.87 } & {\small{}0.78 } & {\small{}0.0016 } & {\small{}3e-06 }\tabularnewline
{\small{}LR } & {\small{}50.37 } & {\small{}29.36 } & {\small{}0.028 } & {\small{}9e-05 } & {\small{}3.19 } & {\small{}0.69 } & {\small{}0.0018 } & {\small{}5e-06 }\tabularnewline
{\small{}OTDR } & \textbf{\small{}22.88}{\small{} } & \textbf{\small{}8.79}{\small{} } & \textbf{\small{}0.00043}{\small{} } & \textbf{\small{} 1e-05}{\small{} } & \textbf{\small{}0.23}{\small{} } & \textbf{\small{}0.28}{\small{} } & \textbf{\small{} 9e-05}{\small{} } & \textbf{\small{}4e-07}{\small{} }\tabularnewline
\hline 
\hline 
{\small{}$\delta=1$ } & {\small{}Mean } & {\small{}STD } & {\small{}Mean } & {\small{}STD } & {\small{}Mean } & {\small{}STD } & {\small{}Mean } & {\small{}STD}\tabularnewline
{\small{}RTR } & \textbf{\small{}110.38}{\small{} } & \textbf{\small{}34.57}{\small{} } & {\small{}0.36 } & {\small{}0.0058 } & \textbf{\small{}1.85}{\small{} } & \textbf{\small{}1.03}{\small{} } & \textbf{\small{}0.0012}{\small{} } & \textbf{\small{} 6e-05}\tabularnewline
{\small{}TDR } & {\small{}552.14 } & {\small{}195.08 } & {\small{}0.04 } & {\small{}0.002 } & {\small{}36.69 } & {\small{}5.99 } & {\small{}0.0013 } & {\small{}7e-05}\tabularnewline
{\small{}VPCR } & {\small{}422.57 } & {\small{}188.02 } & {\small{}2.79 } & {\small{}0.009 } & {\small{}29.49 } & {\small{}7.98 } & {\small{}0.092 } & {\small{}0.0027}\tabularnewline
{\small{}LR } & {\small{}947.11 } & {\small{}202.33 } & {\small{}2.78 } & {\small{}0.0091 } & {\small{}56.93 } & {\small{}8.23 } & {\small{}0.18 } & {\small{}0.00061}\tabularnewline
{\small{}OTDR } & {\small{}311.19 } & {\small{}120.51 } & \textbf{\small{}0.04}{\small{} } & \textbf{\small{}0.002}{\small{} } & {\small{}7.64 } & {\small{}4.00 } & {\small{}0.0015 } & {\small{}0.0001}\tabularnewline
\hline 
\end{tabular}{\small\par}

{\small{}\label{Table: SSE1noniid} }{\small\par}
\end{table}
From Table \ref{Table: SSE1noniid}, we can conclude that the proposed
OTDR outperforms all other methods when the noise level is small and
is only second to RTR when the noise level is large. This superiority
is due to two reasons: 1) OTDR can utilize the tensor structure of
point cloud data, while RTR can capture its smoothness; and 2) OTDR
can simultaneously perform dimension reduction and learn regression
coefficients. To understand the the contribution of each component
(i.e., \emph{one-step approach}, \emph{smoothness, }and \emph{the
tensor structure}) in improving the model accuracy, we take a closer
look at simulation results: a)\textbf{ }\emph{Benefit of the one-step
approach for non-i.i.d noises:} In Case 2 with $\delta=1$, the relative
SSE of TDR is $36.69\%$ compared to $7.64\%$ of OTDR. This indicates
that the advantage of the one-step approach (OTDR) over the two-step
approach (TDR) becomes more pronounced for highly correlated noises.
On the other hand, if the noise is i.i.d, the relative SSE of the
OTDR and TDR are very similar. The reason is that for i.i.d. noises,
most of the variational patterns learned through PCs, directly correlate
with the input variables $X$, not with the noise structure; b)\textbf{
}\emph{Benefit of utilizing the tensor structure}: We compare the
performance of TDR and VPCR, which are two-step approaches. They both
first reduce the dimensions of the point clouds, and then, perform
regression on the low-dimensional features. However, unlike VPCR,
TDR utilizes the tensor structure of the data. If $\theta$ is large,
the relative SSE of TDR is much smaller than that of VPCR, especially
when the noise level $\delta$ is large. For example, for i.i.d noises,
the relative SSE is $0.0013$\% for TDR compared to $0.092\%$ for
VPCR when $\delta=1$; c) \emph{Benefit of capturing smoothness :
}The proposed RTR outperforms other methods when $\delta$ is large.
For example, in Case 1, for non i.i.d case and $\delta=1$, the relative
SSE of RTR is $1.85\%$, much smaller than that of the second best,
OTDR, which is $7.64\%$.

\begin{figure}
\subfloat[RTR]{\centering\includegraphics[width=0.24\linewidth]{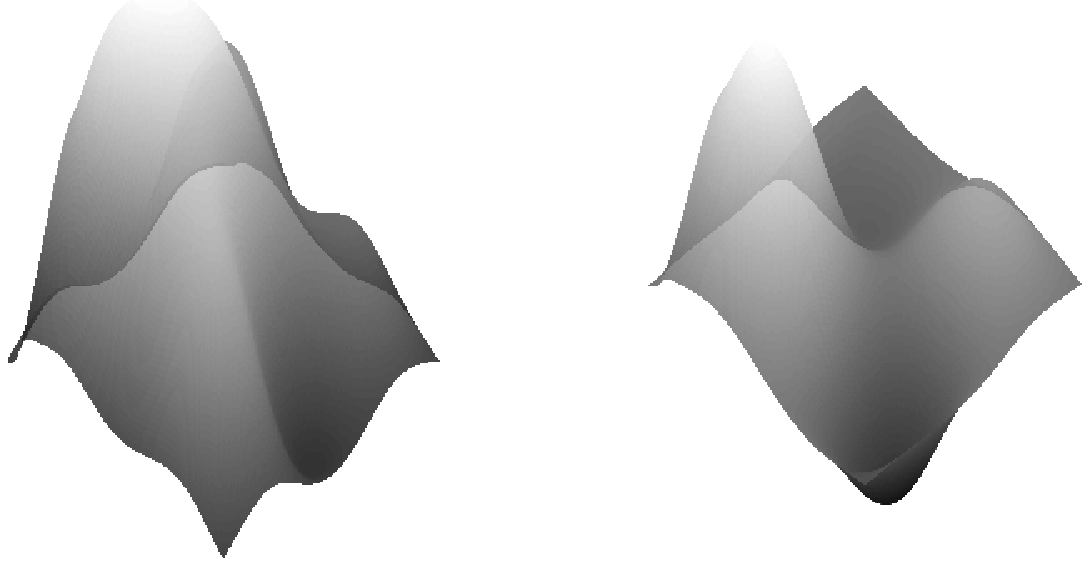}

}\subfloat[OTDR]{\centering\includegraphics[width=0.24\linewidth]{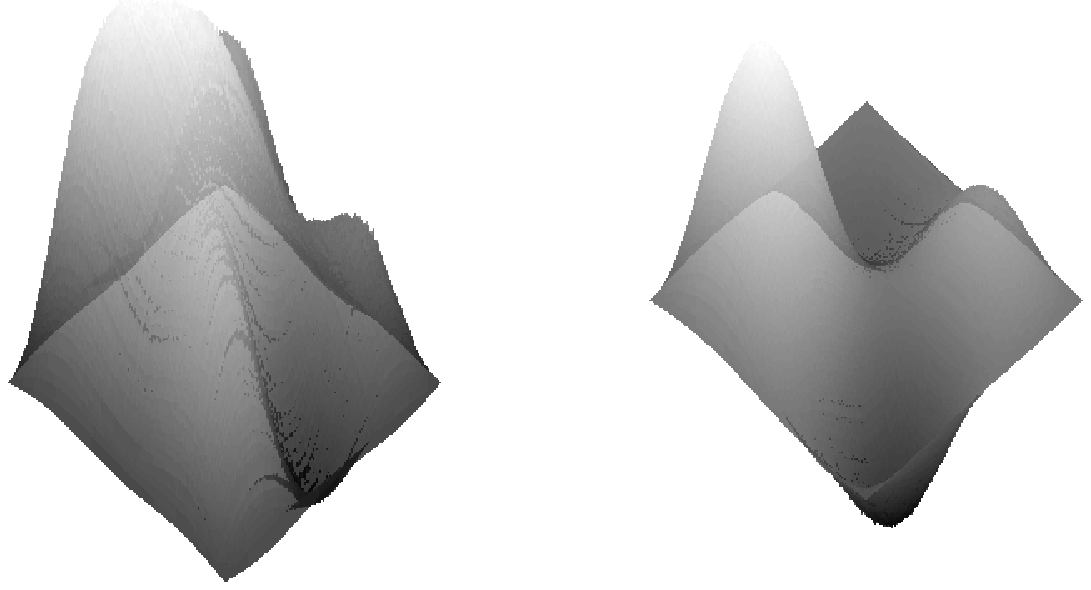}

}\subfloat[TDR]{\centering\includegraphics[width=0.24\linewidth]{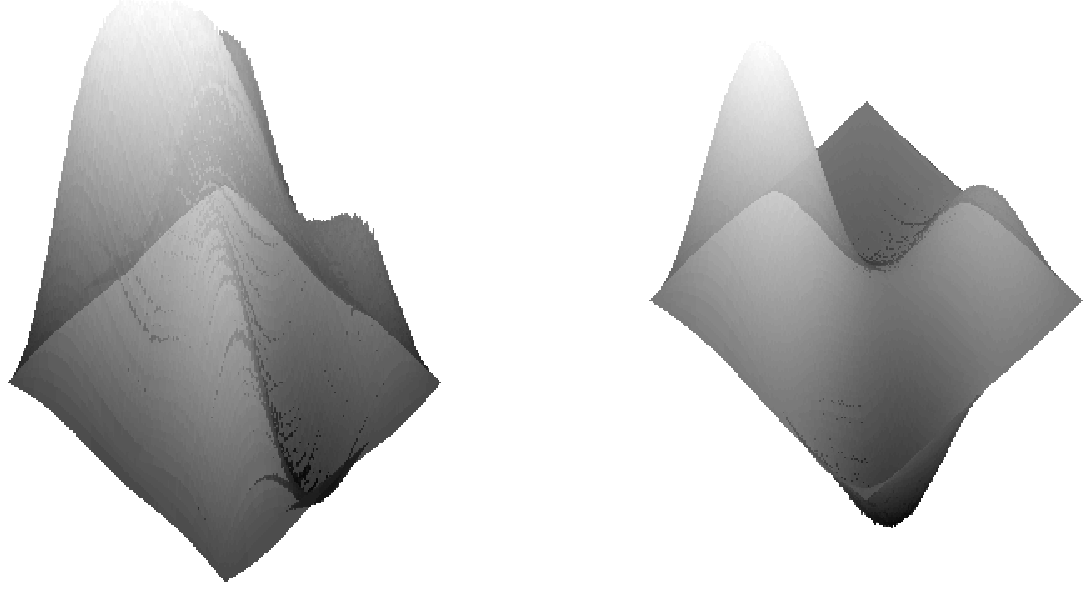}

}\subfloat[VPCR]{\centering\includegraphics[width=0.24\linewidth]{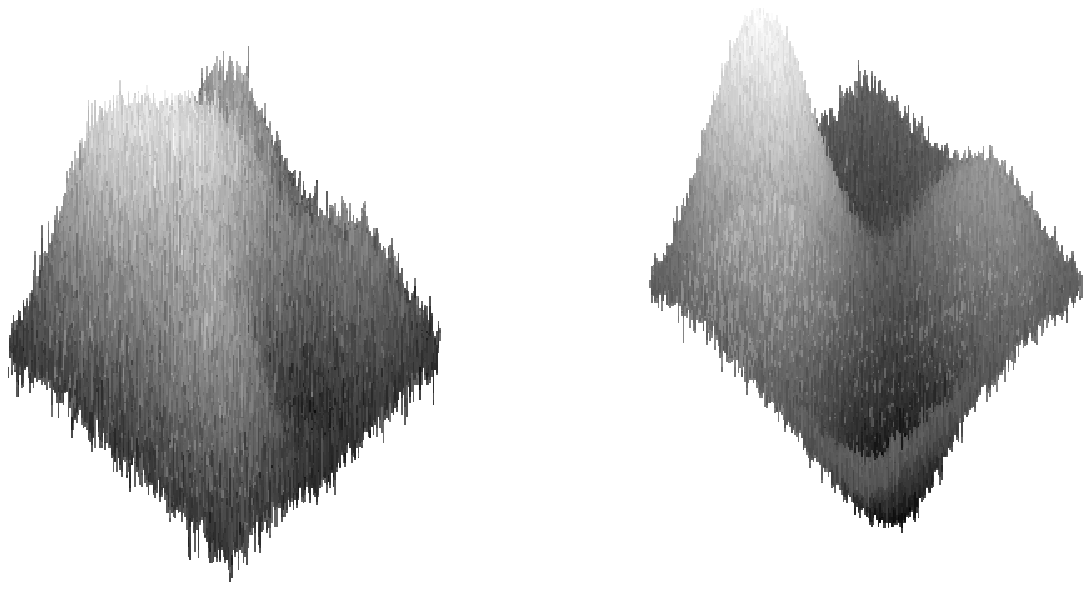}

}

\subfloat[RTR]{\centering\includegraphics[width=0.24\linewidth]{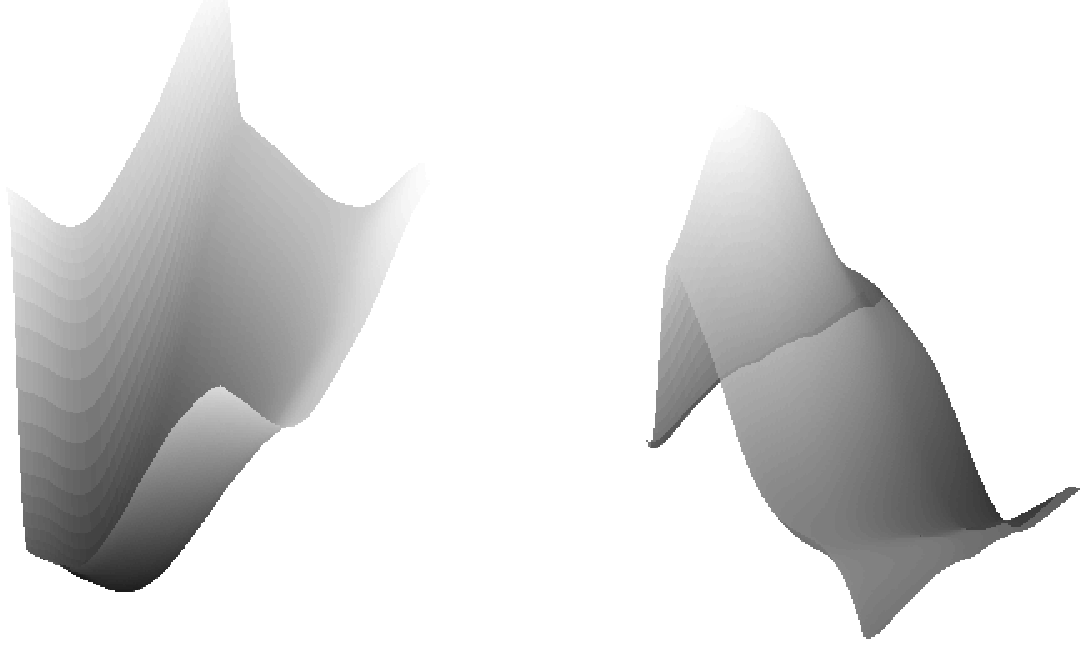}

}\subfloat[OTDR]{\centering\includegraphics[width=0.24\linewidth]{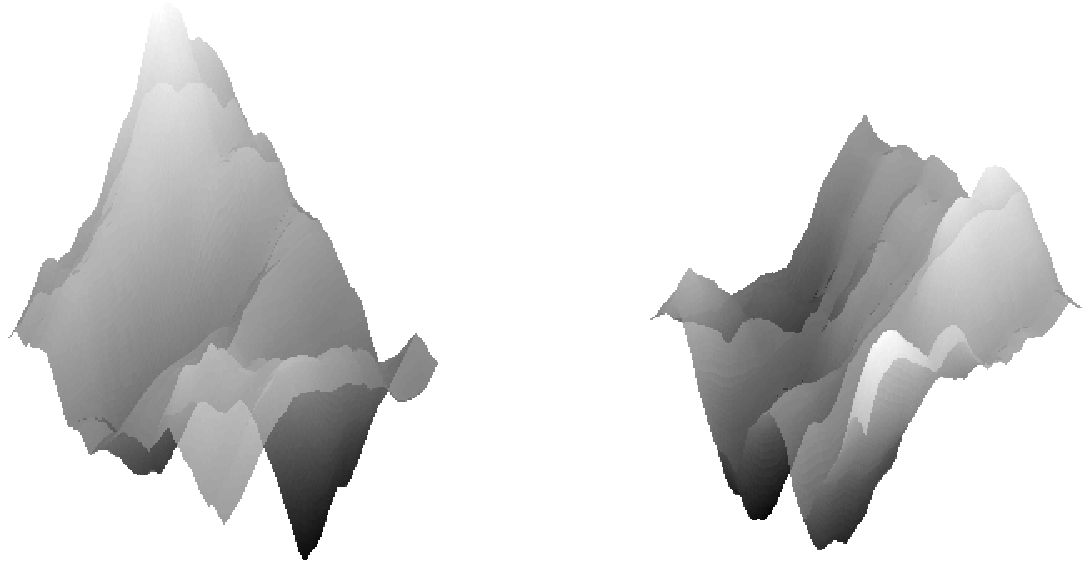}

}\subfloat[TDR]{\centering\includegraphics[width=0.24\linewidth]{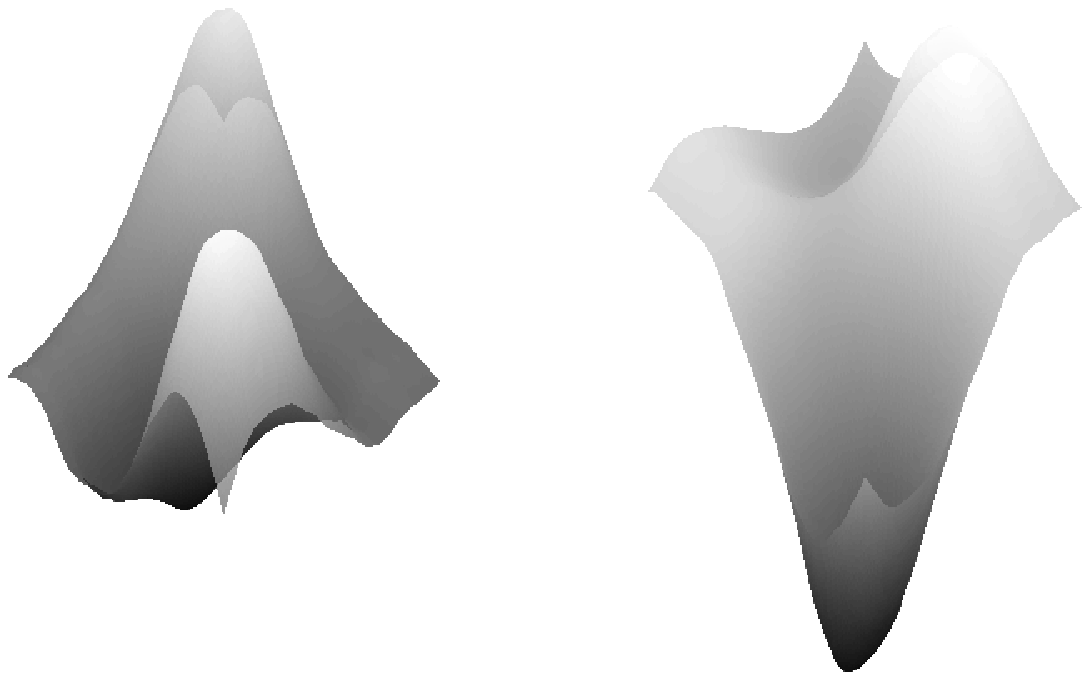}

}\subfloat[VPCR]{\centering\includegraphics[width=0.24\linewidth]{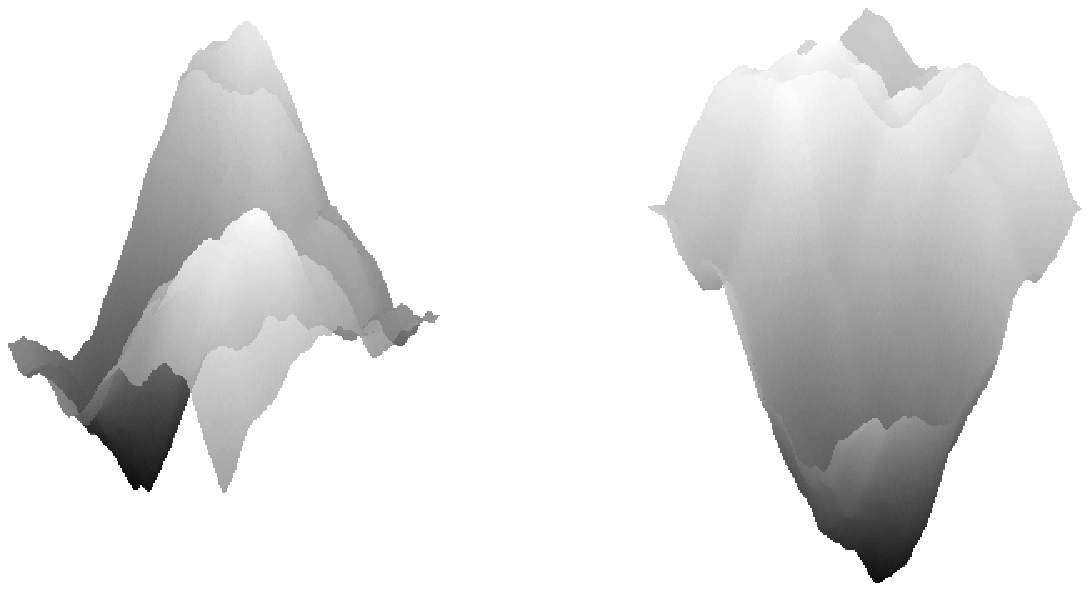}

}

\centering\subfloat[True coefficient]{\centering\includegraphics[width=0.35\linewidth]{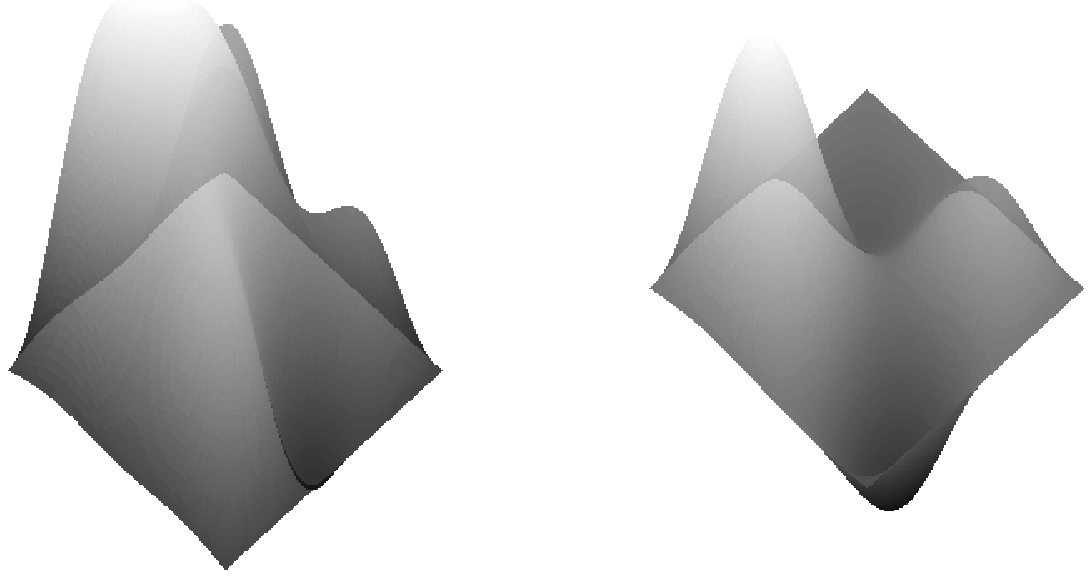}

}

\caption{Estimated and true coefficient for Case 1 with $\delta=0.1$ (1st
row: i.i.d noise; 2nd row non-i.i.d noise with $\theta=10$) }

\label{Fig: case1coef} 
\end{figure}

\begin{figure}
\subfloat[RTR]{\centering\includegraphics[width=0.24\linewidth]{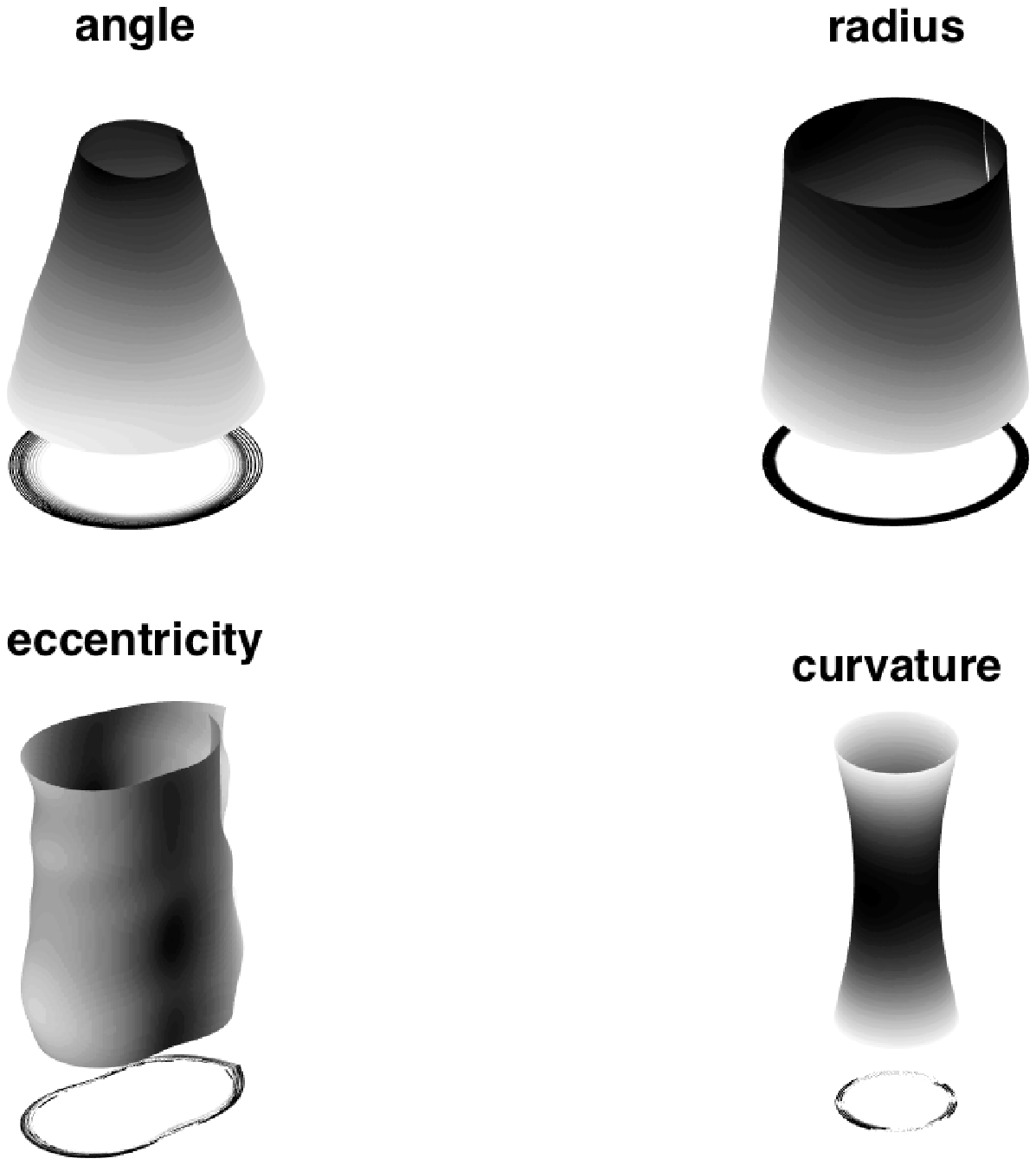}

}\subfloat[OTDR]{\centering\includegraphics[width=0.24\linewidth]{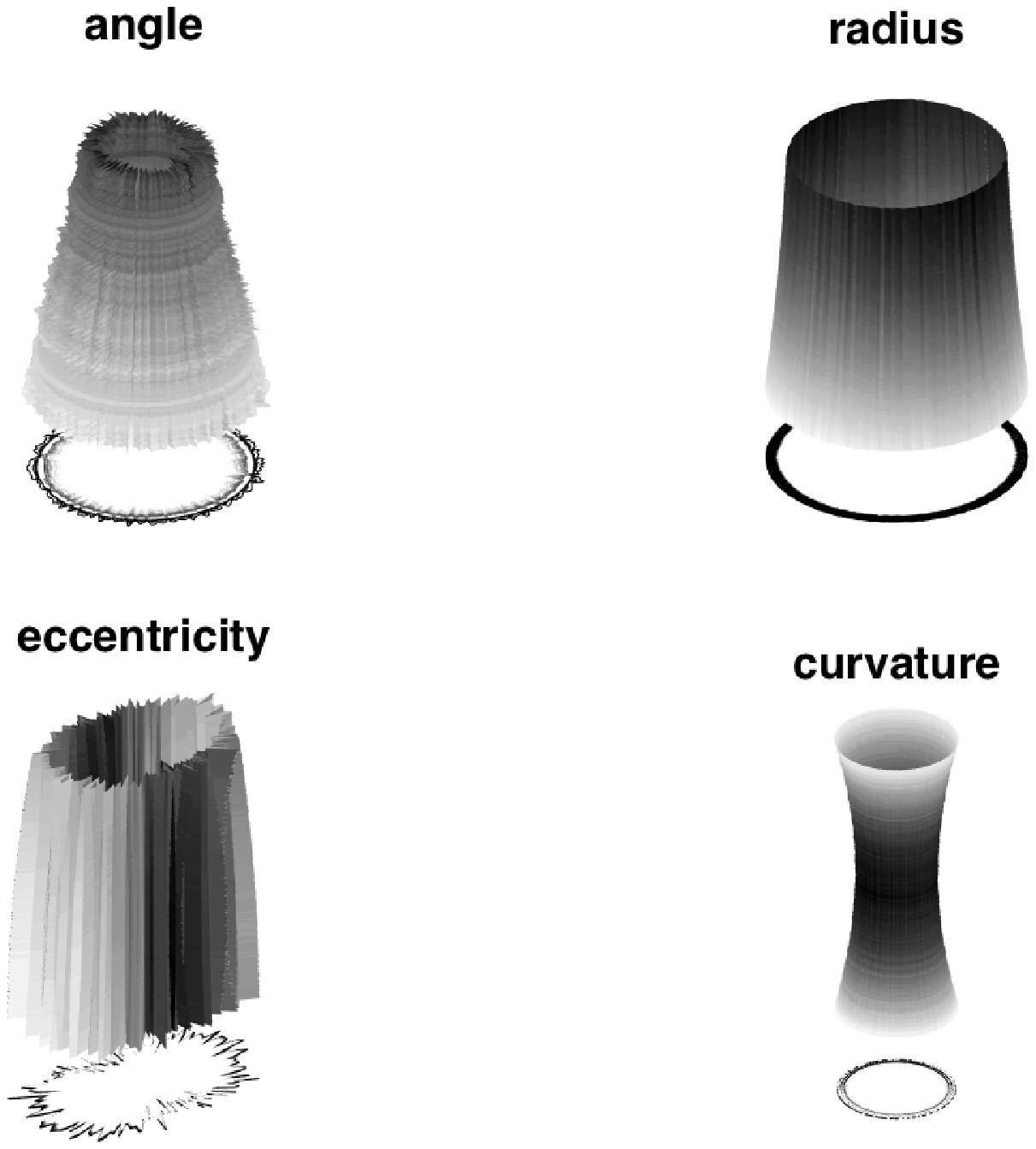}

}\subfloat[TDR]{\centering\includegraphics[width=0.24\linewidth]{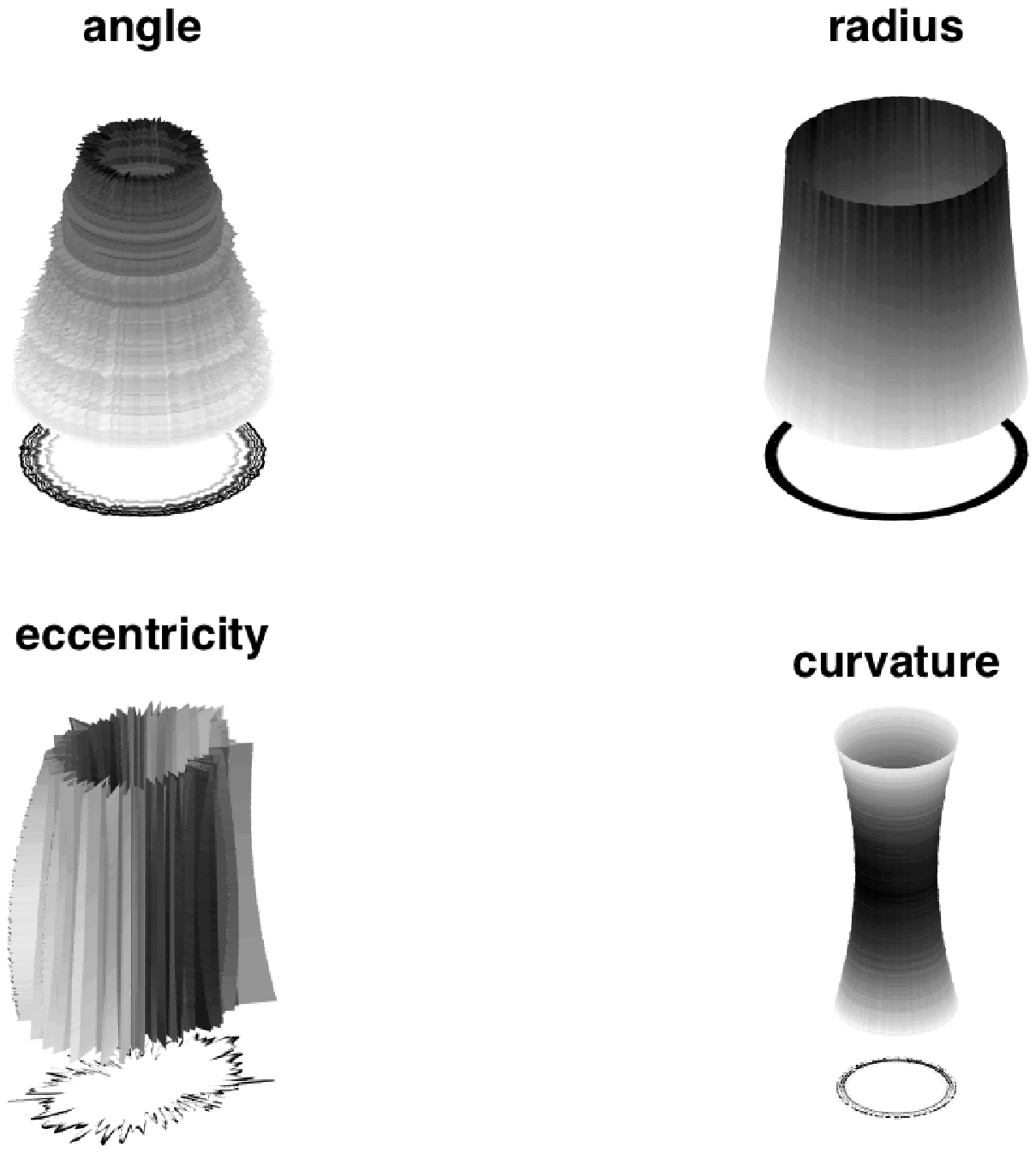}

}\subfloat[VPCR]{\centering\includegraphics[width=0.24\linewidth]{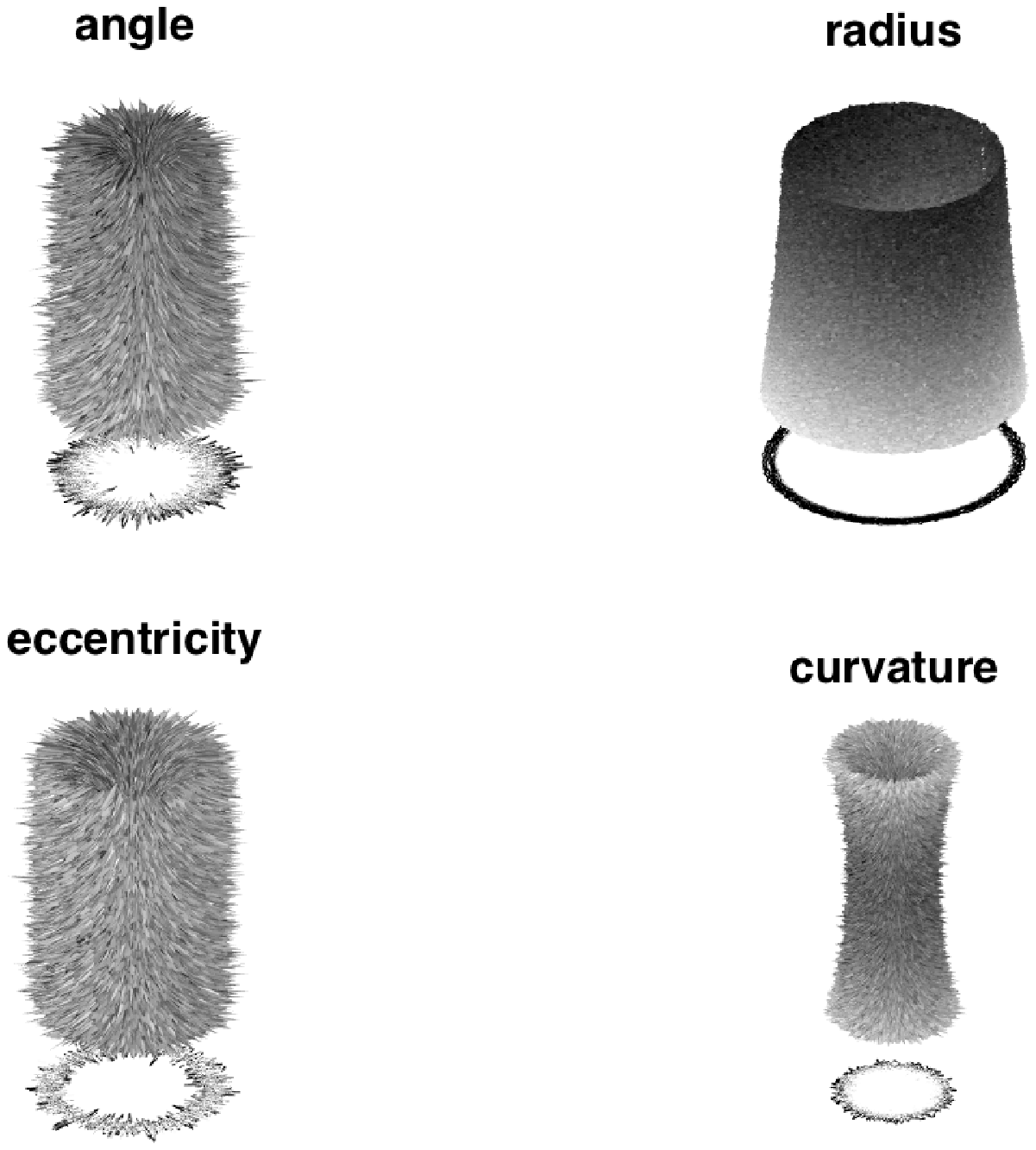}

}

\subfloat[RTR]{\centering\includegraphics[width=0.24\linewidth]{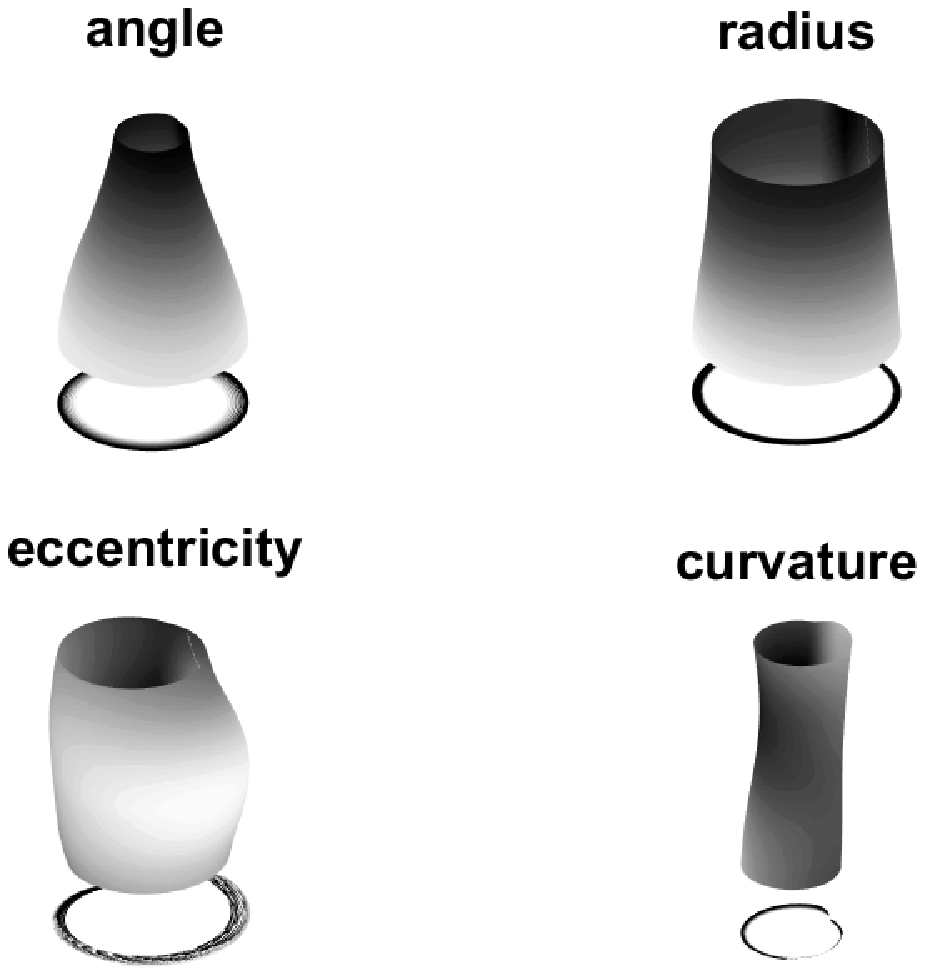}

}\subfloat[OTDR]{\centering\includegraphics[width=0.24\linewidth]{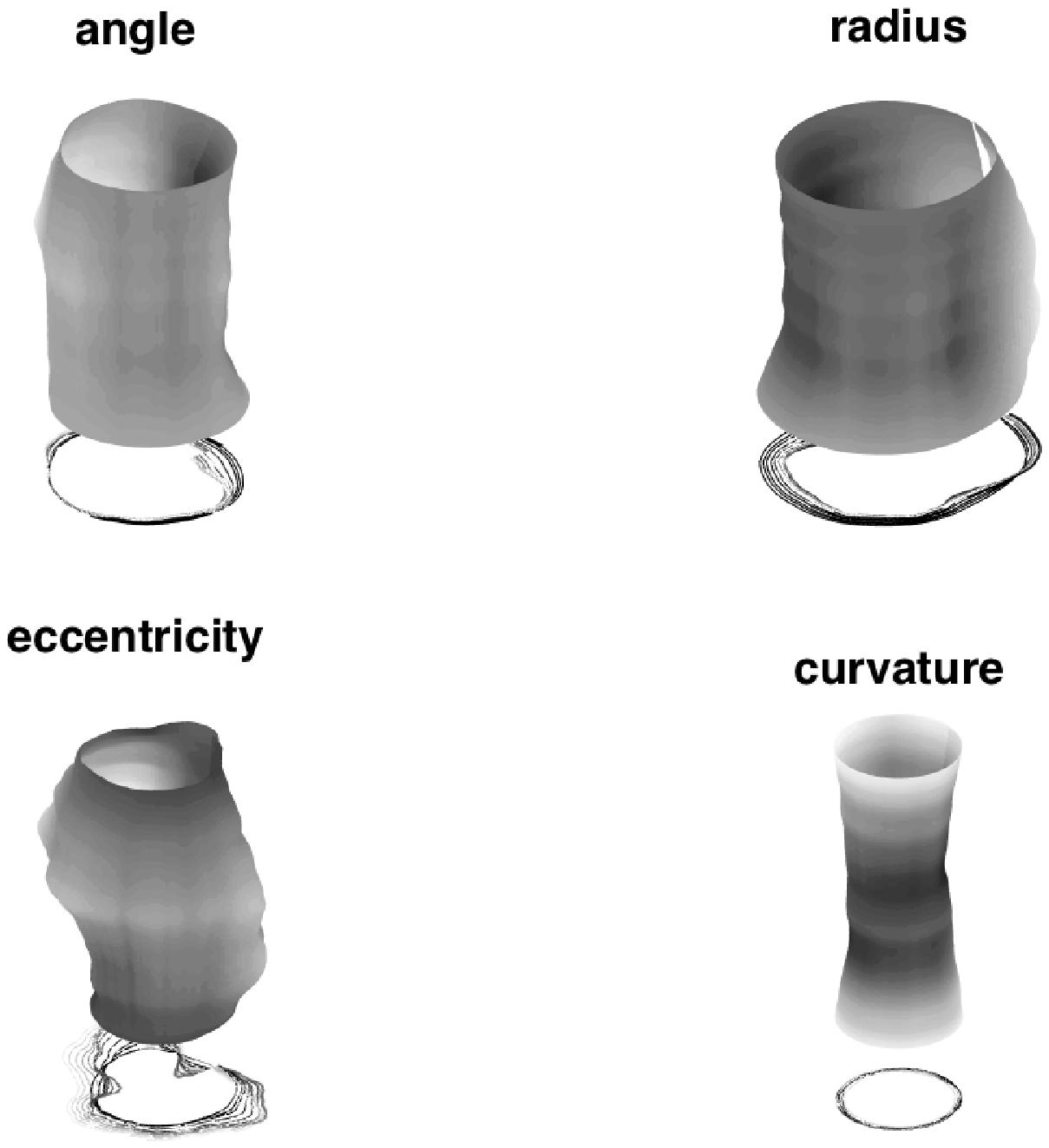}

}\subfloat[TDR]{\centering\includegraphics[width=0.24\linewidth]{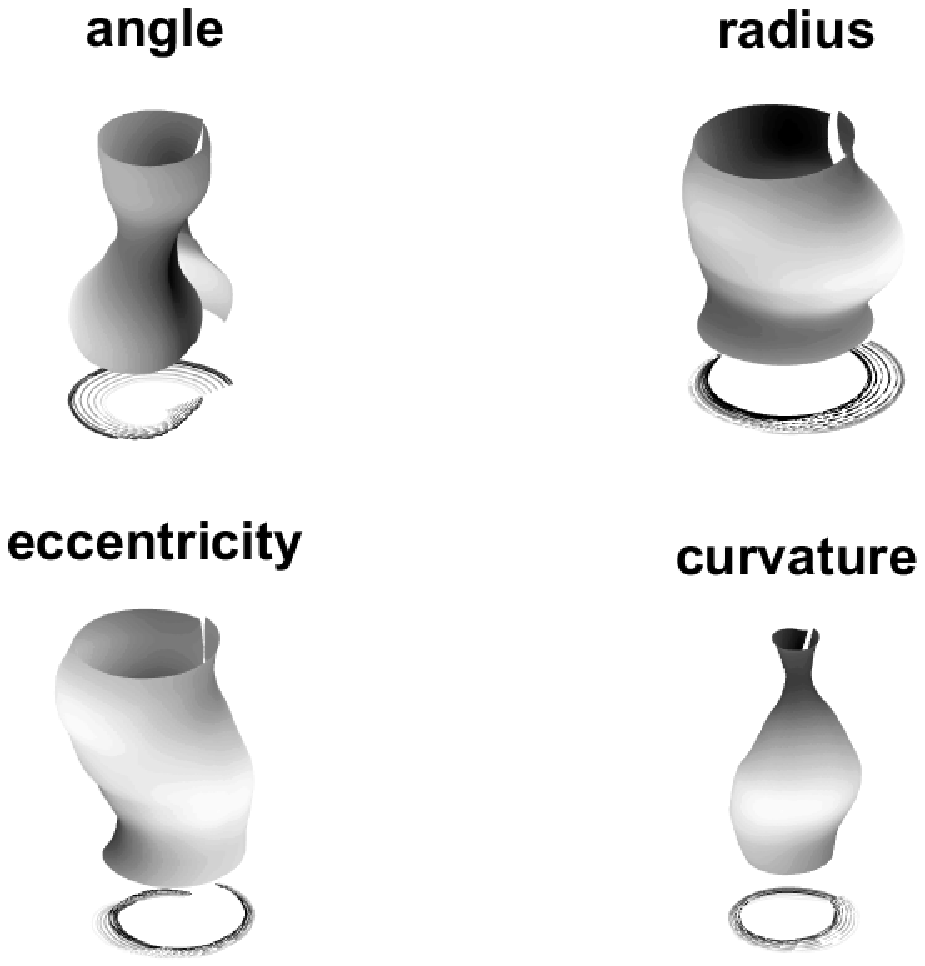}

}\subfloat[VPCR]{\centering\includegraphics[width=0.24\linewidth]{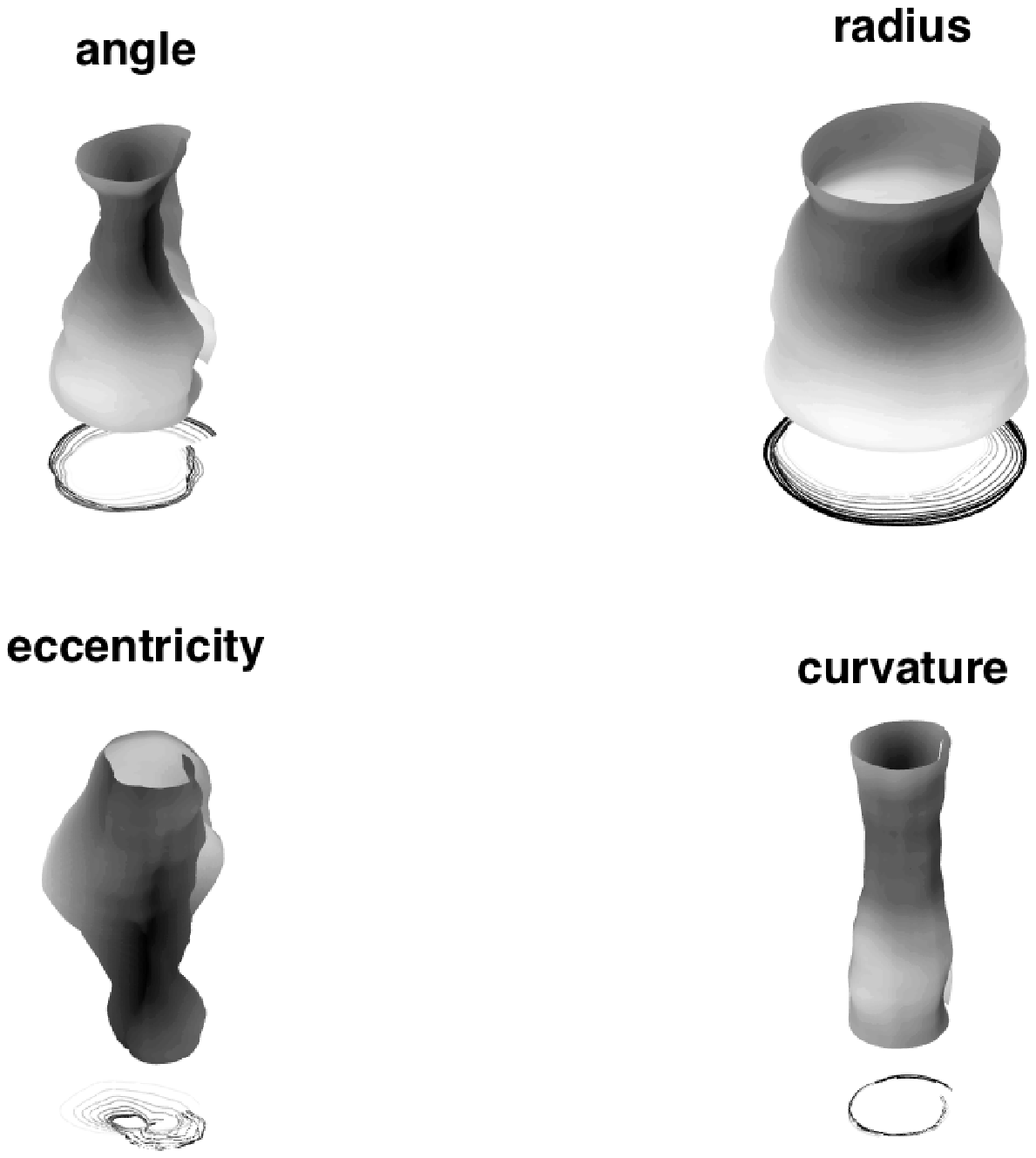}

}

\caption{Estimated and true coefficient for Case 2 with $\delta=0.1$ (1st
row: i.i.d noise; 2nd row non-i.i.d noise with $\theta=10$) }

\label{Fig: case2coef} 
\end{figure}

We then plot the learned coefficient $\mathcal{A}$ for VPCR, OTDR
and RTR for Case 1 and 2 in Figure \ref{Fig: case1coef} and \ref{Fig: case2coef},
respectively. From these plots, we can see that RTR learns a much
smoother coefficient due to the use of smooth basis. However, this
constraint may lead to a larger bias when the noise level is small
for non-i.i.d noises (See Figure \ref{Fig: case1coef} (d) as an example).
Furthermore, although OTDR does not incorporate any prior smoothness,
the learned basis matrices are much smoother than those of VPCR. This
can be seen clearly by comparing Figure \ref{Fig: case1coef} (b),
(f) with (d), (h); and by comparing Figure \ref{Fig: case2coef} (b),
(f) with (d), (h). This is because OTDR utilizes the tensor structure.
Furthermore, in the non i.i.d noise case, all the methods perform
worse than in the case with the i.i.d noise. However, the one-step
approach OTDR still learns more accurate basis than TDR. This can
be seen clearly seen by comparing Figure \ref{Fig: case2coef} (g)
with (h) and by comparing Figure \ref{Fig: case1coef} (g) with (h).


\section{Case Study \label{sec:Case-study}}

In this section, a real case study concerning cylindrical surfaces
obtained by lathe-turning is described and taken as reference in order
to analyze the proposed methods. The study refers to cylinders of
material Ti-6Al-4V, which is a titanium alloy principally used in
the aerospace field because of certain properties (e.g., high specific
strength, corrosion and erosion resistance, high fatigue strength)
that make it the ideal structural material of mechanical components
for both airframes and engines \citep{Peters2003}. While aerospace
applications require significant machining of mechanical components,
titanium and its alloys are generally classified as difficult-to-machine
materials \citep{Pramanik2014,Salman2014}. Combinations of machining
parameters, such as the cutting speed, cutting depth and feed rate,
are varied and optimized in order to improve machinability and thereby
improving part quality.

\subsection{The Reference Experiment \label{sec: experiment}}

Various investigations have been conducted in the literature to achieve
the most favorable cutting conditions of titanium alloys by optimizing
the process parameters \citep{Khanna2015280,Chauhan2012}. In this
paper, the experiment described in \citep{colosimo2016multilinear}
is considered as reference. Bars supplied in $20$mm diameter, were
machined to a final diameter of $16.8$mm by implementing two cutting
steps. During the experiment, the combinations of two parameter values
were varied according to a $3^{2}$ full factorial design. In particular,
with reference to the second cutting step, the cutting speed was set
at $65$, $70$ and $80$ m/min, while cutting depth was set at $0.4$,
$0.8$ and $1.2$ mm. The combinations of parameters used are shown
in Table \ref{table: ProcessVar}, which shows $9$ treatments with
different process variables (each treatment replicated $10$ times).
In order to apply the proposed methods to the experimental data, the
two process variables (cutting depth and cutting speed) were recorded
in the input matrix $\mathbf{X}$ after the normalization (subtract
the mean and divided by the standard deviation).

\begin{table}
\caption{Cutting parameters for $9$ experimental conditions}

\centering%
\begin{tabular}{|c|c|c|}
\hline 
Ex. No  & Depth($mm$)  & Speed($m/min$)\tabularnewline
\hline 
\hline 
1  & 0.4  & 80\tabularnewline
\hline 
2  & 0.4  & 70\tabularnewline
\hline 
3  & 0.4  & 65\tabularnewline
\hline 
4  & 0.8  & 80\tabularnewline
\hline 
5  & 0.8  & 70\tabularnewline
\hline 
6  & 0.8  & 65\tabularnewline
\hline 
7  & 1.2  & 80\tabularnewline
\hline 
8  & 1.2  & 70\tabularnewline
\hline 
9  & 1.2  & 65\tabularnewline
\hline 
\end{tabular}

\label{table: ProcessVar} 
\end{table}
The $90$ cylindrical surfaces were measured with a CMM using a touch
trigger probe with a tip stylus of $0.5$ mm radius. The measurements
were taken in $42$ mm along the bar length direction with $210$
cross-sections. Each cross-section was measured with $64$ generatrices.
A set of $210\times64$ points, equally distributed on the cylindrical
surface, was measured for each sample $\mathbf{Y}_{i},i=1,\cdots,90$.
All the samples were aligned by rotation \citep{silverman1995incorporating}.
The final surface data was computed as deviations of the measured
radii from a reference cylinder, which was computed using a least-square
approach. By subtracting the radius of the substitute geometry, the
final set of measurements consists of a set of radial deviations from
a perfect cylinder, measured at each position.

The examples of the cylindrical surfaces are shown in Figure \ref{Fig: surface},
which clearly shows that the shape of the cylinder is influenced by
both the cutting speed and cutting depth. In particular, a \emph{taper}
axial form error \citep{Henke1999273} is the most evident form error
for cylindrical surfaces. This was mainly due to deflection of the
workpiece during turning operations. The degree of the deflection
at cutting location varies and depends on how far the cutting tool
is from the supporters of the chuck \citep{zhang2005cylinder}.

\subsection{Surface Roughness \label{sec: roughtness}}

While surface shape represents the overall geometry of the area of
interest, surface roughness is a measurement of the surface finish
at a lower scale (surface texture). Surface roughness is commonly
characterized with a quantitative calculation, expressed as a single
numeric parameter of the roughness surface, which is obtained from
a high-pass filtering of the measured surface after the shortest wavelength
components are removed. In the reference case study, the measured
surface is obtained from scanning the actual surface with a probe
which mechanically filters this data due to the CMM tip radius (0.5
mm). Given the CMM measurements in the experiments, the variance of
residuals after modeling the shape of cylindrical items is assumed
as the quantitative parameter related to surface roughness.

In machining the titanium alloy, several surface roughness prediction
models - in terms of the cutting speed, feed rate and cutting depth
- have been reported in the literature (e.g., \citet{Ramesh20121266}).
In general, it has been found that cutting depth is the most significant
machining parameter. In the experiment of \citep{colosimo2016multilinear}
an unequal residual variance, caused by different process variables,
was observed.

\subsection{Handling Unequal Variances of Residuals \label{sec: unequal}}

In order to model both the cylindrical mean shape and the residuals
with unequal variance (unequal surface roughness) caused by the different
process variables, we combine the framework proposed by \citep{western2009variance}
with our proposed tensor regression model in Section \ref{sec: unequal}.

To model the unequal variances of residuals as a function of the process
variables, we assume that the noise $E_{i}\sim N(0,\sigma_{i}^{2})$,
where $\log\sigma_{i}^{2}=\mathbf{x}_{i}'\boldsymbol{\gamma}+\gamma_{0}$.
Therefore, by combining it with the tensor regression model in \eqref{eq: matrixreg},
the maximum likelihood estimation (MLE) method can be used to estimate
the parameters $\boldsymbol{\gamma}$, $\gamma_{0}$ and $\mathcal{A}$.
The likelihood function (for the i.i.d noise) is given by $L(\beta,\boldsymbol{\gamma};y_{i})=-\frac{1}{2}(\sum_{i}I_{1}I_{2}\log(\sigma_{i}^{2})+\sum_{i}\frac{\|\mathbf{Y}_{i}-\bar{\mathbf{Y}}-\mathcal{A}\times_{3}\mathbf{X}_{i}\|^{2}}{\sigma_{i}^{2}})$,
which can be maximized by iteratively updating $\boldsymbol{\gamma}$
and $\mathcal{A}$ until convergence according to the following procedure:
1) For the fixed $\boldsymbol{\gamma}$ and $\gamma_{0}$, perform
transformations given by $\mathbf{Y}_{i}^{0}=\frac{\mathbf{Y}_{i}-\bar{\mathbf{Y}}}{\sigma_{i}}$,
$\mathbf{X}_{i}^{0}=\frac{\mathbf{X}_{i}}{\sigma_{i}}$, where $\sigma_{i}^{2}=\exp(\mathbf{x}_{i}'\boldsymbol{\gamma}+\gamma_{0})$.
The resulting MLE can be obtained by the proposed tensor regression
methods introduced in Section \eqref{sec:Tensor-Regression-Model}.
2) For fixed $\mathcal{A}$, MLE reduces to gamma regression with
$\log$ link on the Residual Mean Squares Error (RMSE), i.e., $\frac{1}{I_{1}I_{2}}\|\hat{\mathbf{E}}_{i}\|^{2}$,
where $\hat{\mathbf{E}}_{i}=\mathbf{Y}_{i}-\mathcal{A}\times_{3}\mathbf{X}_{i}$.

\begin{figure}
\centering\includegraphics[width=0.8\linewidth]{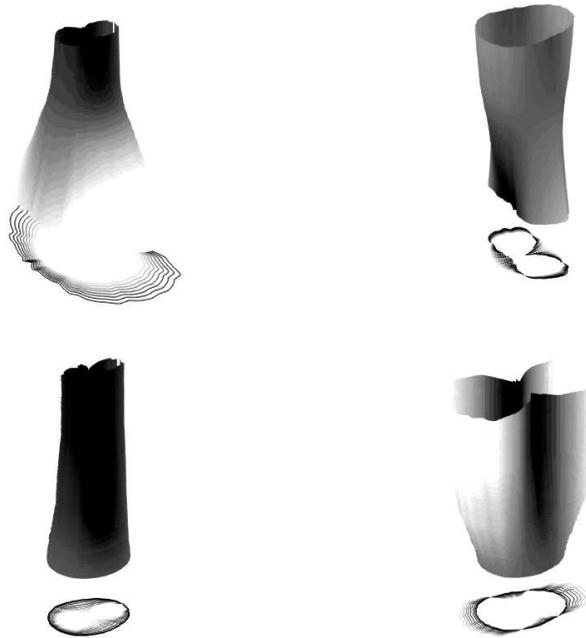}

\caption{Eigentensors learned from OTDR with regularized Tucker decomposition}

\label{Fig: casetuckerbasis} 
\end{figure}

We then apply OTDR on these cylindrical surfaces to map the relationship
of the mean shape and residual variance with process variables. The
first nine ($3\times3$) eigentensors of OTDR are extracted and shown
in Figure \ref{Fig: casetuckerbasis}.

\begin{figure}
\subfloat[Tensor regression coefficient $\mathcal{A}$]{\centering\includegraphics[width=0.45\linewidth]{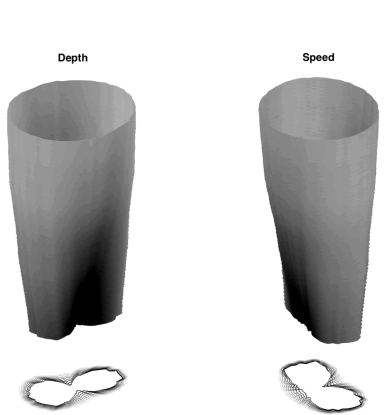}

\label{Fig: casecoef}

}\hfill{}\subfloat[Residual Mean of Square Error (RMSE) and fitted $\hat{\sigma}^{2}$
via gamma regression]{\centering\includegraphics[width=0.45\linewidth]{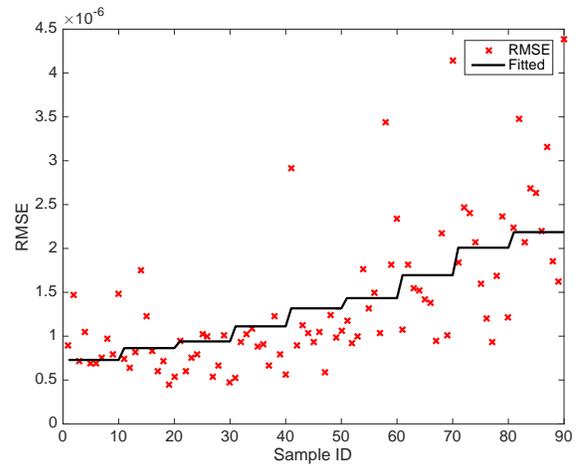}

\label{Fig: caseRSS}

}

\caption{Result of tensor regression via OTDR}

\label{Fig: casetensorreg} 
\end{figure}

From Figure \ref{Fig: casetuckerbasis}, it can be observed that the
shape of eigentensors can be interpreted as the combination of both
bi-lobed and three-lobed contours along the radial direction, to conical
shapes along the axial direction. This result is consistent with that
reported in the literature where a conical shape along the vertical
(referred to as \emph{taper} error) was defined as a ``dominant''
axial form error of manufactured cylindrical surfaces. Similarly,
radial form errors are often described as bi-lobed (oval) and three-lobed
(three-cornered) contours, which are typical harmonics that characterize
the cross-section profiles of cylinders obtained by lathe-turning
\citep{Henke1999273}. The estimated coefficients, $\hat{\mathcal{A}}$,
are also shown in Figure \ref{Fig: casecoef}. A visual inspection
of this figure shows that the systematic shape characterizing the
tensor regression coefficients relates to the cutting depth and speed
parameters, i.e., a conical shape whose inferior portion assumes a
bi-lobed contour. Again, the result is consistent with that reported
in the literature where a conical shape is a common axial form error
of manufactured cylindrical surfaces, and a bi-lobed contour is a
typical harmonic that characterize the cross-section profiles of cylinders
obtained by lathe-turning \citep{Henke1999273}.

The RMSE and the fitted $\sigma^{2}$ of the $90$ samples via the
gamma regression are shown in Figure \ref{Fig: caseRSS}. It is clear
that the proposed framework is able to account for unequal variances
under the $9$ different input settings. The gamma regression coefficients
of the RMSE are reported in Table \ref{Table: TableCoef}. From this
table, we can conclude that if the cutting depth increases or cutting
speed decreases, the variance of residuals will also increase. Moreover,
for the variance of residuals (surface roughness), the effect of the
cutting depth is much more significant than the that of the cutting
speed. These findings are consistent with engineering principles.

\begin{table}
\caption{Gamma regression of $\|\hat{E}_{i}\|^{2}$ }

\centering%
\begin{tabular}{|c|c|c|c|c|c|}
\hline 
\multicolumn{2}{|c|}{} & Estimate  & SE  & t-Stat  & p-Value\tabularnewline
\hline 
\hline 
$\gamma_{0}$  & Intercept  & $-13.5681$  & $0.0455$  & $-298.12$  & $1e-133$\tabularnewline
\hline 
\multirow{2}{*}{$\boldsymbol{\gamma}$}  & depth  & $0.3654$  & $0.0483$  & $7.568$  & $3.7e-11$\tabularnewline
\hline 
 & speed  & $-0.1121$  & $0.0483$  & $-2.322$  & $0.02$\tabularnewline
\hline 
\end{tabular}

\label{Table: TableCoef} 
\end{table}

\subsection{Process Optimization}

\begin{figure}
\subfloat[Relationship of optimal cutting depth and $\sigma_{0}$]{\centering\includegraphics[height=0.3\paperheight]{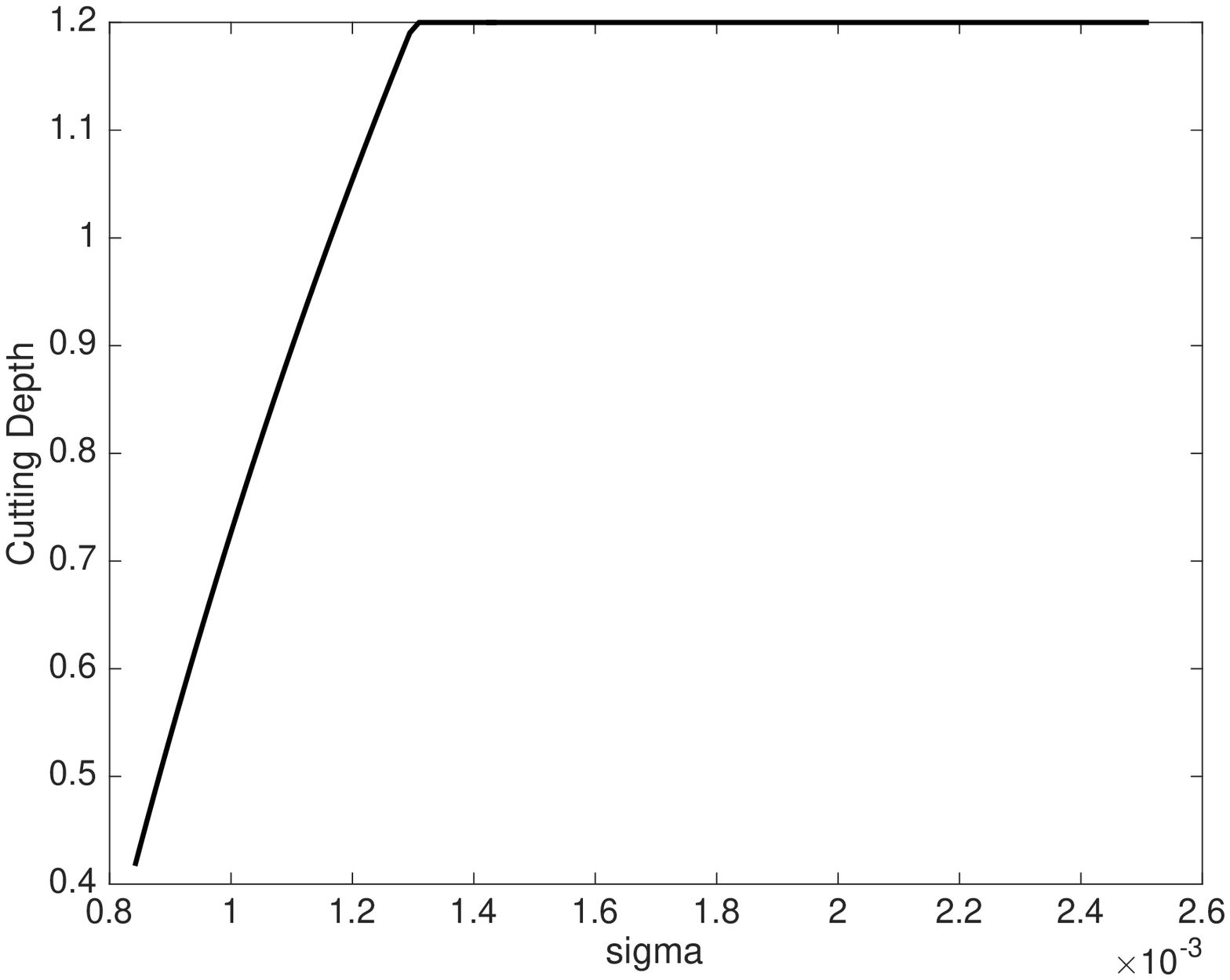}

\label{Fig: cuttingdepth}

}\hfill{}\subfloat[Simulated cylinder under the optimal setting. Scale 250:1]{\centering\includegraphics[height=0.3\paperheight]{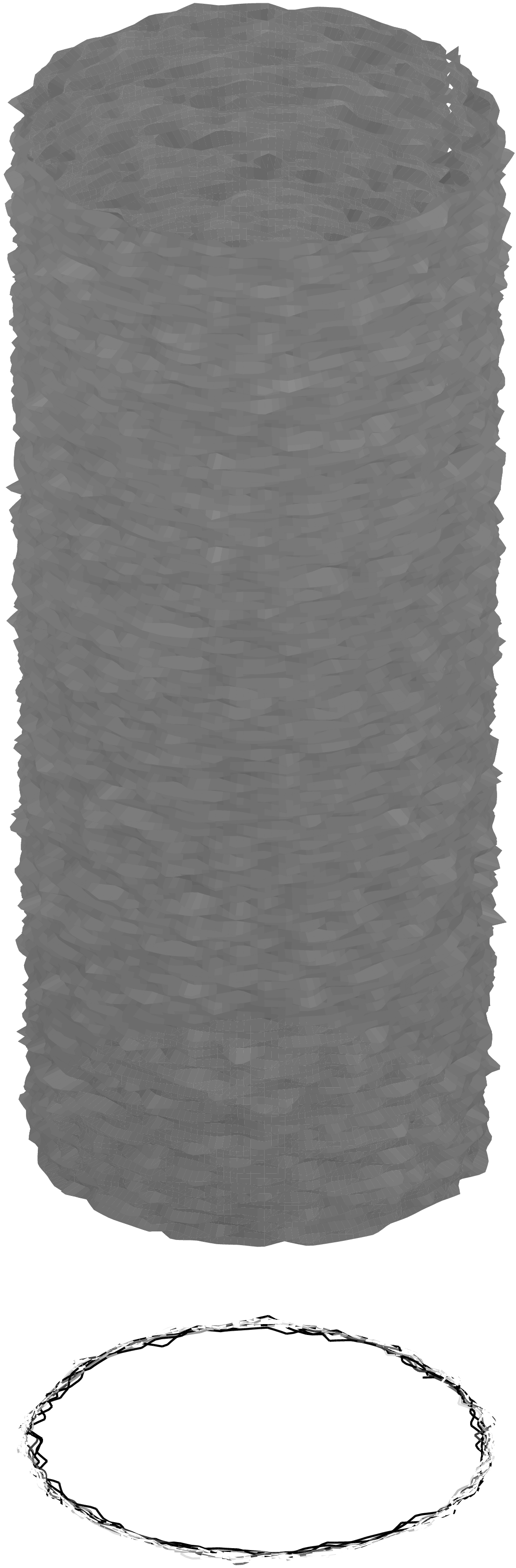}

\label{Fig: PredictedCylinder}

}

\caption{Optimal Settings for different $\sigma$}
\end{figure}

The estimated tensor regression model can also provide useful information
to optimize the process settings (cutting depth and cutting speed)
for better product quality. In this turning process, the goal is to
produce a cylindrical surface with a uniform radius $r_{t}=16.8\mathrm{mm}$.
Therefore, the objective function is defined as the sum of squared
differences of the produced mean shape and the uniform cylinder with
radius $r_{t}$. Furthermore, we require the produced variance of
residuals, $\sigma$ , to be smaller than a certain threshold, $\sigma_{0}$.
Finally, the process variables are typically constrained in a certain
range defined by $\mathbf{l}\leq\mathbf{x}\leq\mathbf{u}$ due to
the physical constraints of the machine. Therefore, the following
convex optimization model can be used for optimizing the turning process:
\[
\min_{\mathbf{x}}\|\bar{\mathbf{Y}}+\hat{\mathcal{A}}\times_{3}\mathbf{x}-r_{t}\|_{F}^{2}\quad s.t.\sigma\leq\sigma_{0},\mathbf{l}\leq\mathbf{x}\leq\mathbf{u}.
\]
It is straightforward to show that this optimization problem can be
reformulated to a quadratic programming (QP) model with linear constraints
as 
\[
\min_{\mathbf{x}}\mathbf{x}^{T}\mathbf{A}_{(3)}^{T}\mathbf{A}_{(3)}\mathbf{x}+2\mathbf{x}^{T}\mathbf{A}_{(3)}^{T}(\mathrm{vec}(\bar{\mathbf{Y}})-r_{t})\quad s.t.\boldsymbol{\gamma}'\mathbf{x}\leq\log(\sigma_{0}^{2})-\gamma_{0},\mathbf{l}\leq\mathbf{x}\leq\mathbf{u}.
\]
Since the problem is convex, it can be solved via a standard quadratic
programming solver. For example, if $\sigma_{0}=0.0001$ and process
variables are limited to the range defined by the design matrix in
Table \ref{table: ProcessVar}, the optimal cutting speed and cutting
depth are computed as $80\mathrm{m/min}$ and $0.7264\mathrm{mm}$,
respectively. Under this setting, we simulate the produced cylindrical
surfaces as shown in Figure \ref{Fig: PredictedCylinder} by combining
both the predicted surface $\hat{\mathbf{Y}}=\bar{\mathbf{Y}}+\hat{\mathcal{A}}\times_{3}\mathbf{x}$
and generated noises from the normal distribution with the estimated
standard deviation by $\hat{\sigma}=\exp(\frac{1}{2}(\gamma_{0}+\boldsymbol{\boldsymbol{\gamma}}'\hat{\mathbf{x}}))$.
It is clear that the produced cylindrical surfaces under this optimal
setting is closer to the uniform cylinder compared than other input
settings as shown in Figure \ref{Fig: surface}. To show the optimal
setting for different levels of the $\sigma_{0}$, we plot the relationship
between the optimal cutting depth and the right hand side of the roughness
constraint, $\sigma_{0}$ in Figure \ref{Fig: cuttingdepth}. It is
worth noting that for all $\sigma_{0}$ larger than $0.8\times10^{-3}$,
the optimal cutting speed stays at the upper bound $80\mathrm{m/min}$,
since the higher cutting speed helps to reduce the variance of residuals
and make the overall shape more uniform. For $\sigma_{0}\leq0.8\times10^{-3}$,
the problem is not feasible, since it is not possible to reduce the
surface variation to a level smaller than $0.8\times10^{-3}$. For
the case where $1.3\times10^{-3}\leq\sigma_{0}\leq0.8\times10^{-3}$,
the higher values of cutting depth results in more uniform cylinder.
However, for $\sigma_{0}\geq1.3\times10^{-3}$, when the variance
of residuals constraint is not the bottle neck, the optimal cutting
depth stays at $1.2$ mm, which is determined by the optimal mean
shape requirement.

\section{Conclusion\label{sec:Conclusion}}

Point cloud modeling is an important research area with various applications
especially in modern manufacturing due to the ease of accessibility
to 3D scanning tools and the need for accurate shape modeling and
control. As most structured point clouds can be represented in a tensor
form in a certain coordinate system, in this paper, we proposed to
use tensor regression to link the shape of point cloud data with some
scalar process variables. However, since the dimensionality of the
tensor coefficients in the regression is too high, we suggested to
reduce the dimensionality by expanding the tensors on a few sets of
basis. To determine the basis, two strategies (i.e. RTR and OTDR)
were developed. In the simulation study, we showed both RTR and OTDR
outperform the existing vector-based techniques. Finally, the proposed
methods were applied to a real case study of the point cloud modeling
in a turning process. To model unequal variances due to the roughness,
an iterative algorithm for maximum likelihood estimation was proposed
which combined the proposed tensor regression model with gamma regression.
The results indicated that our methods were capable of identifying
the main variational pattern caused by the input variables. We also
demonstrated that how this model could be used to find the optimal
setting of process variables.

There are several potential research directions to be investigated.
One direction is to extend this method to non-smooth point clouds
with abrupt changes in surface. Another direction is to develop a
modeling approach for unstructured point clouds. Also, the tensor
regression problem where both input and response variables are high-order
tensors is an interesting, yet challenging problem for future research.

\appendix

\section{Optimizing the Likelihood Function Without Basis \label{sec:Lieklihoodderive}}

The tensor coefficients can be estimated by minimizing the negative
likelihood function $\mathbf{a}$, i.e., $\hat{\mathbf{a}}=\min_{\mathbf{a}}(\mathbf{y}-(\mathbf{X}\otimes I\otimes I)\mathbf{a})^{T}(\Sigma_{3}\otimes\Sigma_{2}\otimes\Sigma_{1})^{-1}(\mathbf{y}-(\mathbf{X}\otimes I\otimes I)\mathbf{a}).$
This can be solved by $\hat{\mathbf{a}}=((\mathbf{X}^{T}\Sigma_{3}^{-1}\mathbf{X})^{-1}\mathbf{X}^{T}\Sigma_{3}^{-1}\otimes I\otimes I)\mathbf{y}$,
where $\mathbf{y}=\mathrm{vec}(\mathcal{Y})$ and $\mathbf{a}=\mathrm{vec}(\mathcal{A})$
are the vectorized $\mathcal{Y}$ and $\mathcal{A}$, respectively.
This is equivalent to the tensor format equation, which is $\hat{\mathcal{A}}=\mathcal{Y}\times_{3}(\mathbf{X}^{T}\Sigma_{3}^{-1}\mathbf{X})^{-1}\mathbf{X}^{T}\Sigma_{3}^{-1}$.

\section{Tucker Decomposition Regression \label{sec:Tucker-Decomposition-Regression}}

Principal component analysis (PCA) \citep{jolliffe2002principal}
has been widely used because of its ability to reduce the dimensionality
of high-dimensional data. However, as pointed out by \citet{yan2015image},
applying PCA directly on tensor data requires unfolding the original
tensor into a long vector, which may result in the loss of structural
information of the original tensor. To address this issue, tensor
decomposition techniques such as Tucker decomposition \citep{tucker1966some}
have been proposed and widely applied in image denoising, image monitoring,
tensor completion, etc. Tucker decomposition aims to find a set of
orthogonal transformation matrices $\mathbf{U}=\{\mathbf{U}^{(k)}\in\mathbb{R}^{I_{k}\times P_{k}};\mathbf{U}^{(k)^{T}}\mathbf{U}^{(k)}=\mathbf{I}_{P_{k}},P_{k}<I_{k},k=1,2\}$
such that it can best represents the original data $\mathcal{Y}$,
where $\mathbf{I}_{P_{k}}$ represents the identity matrix of size
$P_{k}\times P_{k}$. That is,

\begin{equation}
\{\mathcal{\hat{S}},\hat{\mathbf{U}}^{(1)},\hat{\mathbf{U}}^{(2)}\}=\mathop{\mathrm{argmin}}_{\mathcal{S},\mathbf{U}^{(1)},\mathbf{U}^{(2)}}\|\mathcal{Y}-\mathcal{S}\times_{1}\mathbf{U}^{(1)}\times_{2}\mathbf{U}^{(2)}\|_{F}^{2}.\label{eq: Tucker-1}
\end{equation}
$\mathcal{\hat{S}}$ is the core tensor and can be obtained by 
\begin{equation}
\mathcal{\hat{S}}=\mathcal{Y}\times_{1}\hat{\mathbf{U}}^{(1)^{T}}\times_{2}\hat{\mathbf{U}}^{(2)^{T}}.\label{eq: coreTensor-1}
\end{equation}
\citet{yan2015image} showed that \eqref{eq: Tucker-1} is equivalent
to maximizing the variation of the projected low-dimensional tensor,
known as multi-linear principal component analysis (MPCA) method proposed
in \citep{lu2008mpca}. Therefore, for finding the basis matrix, one
can solve the following optimization problem: 
\begin{equation}
\{\mathbf{\hat{U}}^{(1)},\mathbf{\hat{U}}^{(2)}\}=\mathop{\mathrm{argmax}}_{\mathbf{U}^{(1)},\mathbf{U}^{(2)}}\|\mathcal{Y}\times_{1}\mathbf{U}^{(1)^{T}}\times_{2}\mathbf{U}^{(2)^{T}}\|_{F}^{2}.\label{eq: MPCA-1}
\end{equation}
$\hat{\mathcal{B}}$ can then be computed by \eqref{eq: coefReg}
and $\hat{\mathcal{A}}=\hat{\mathcal{B}}\times_{1}\hat{\mathbf{U}}^{(1)}\times_{2}\hat{\mathbf{U}}^{(2)}$.

\section{\label{sec: scorereg}The Proof of Proposition \ref{prop: DataReg}}
\begin{proof}
The likelihood function can be minimized by: 
\begin{align*}
\hat{\boldsymbol{\beta}} & =\arg\min_{\boldsymbol{\beta}}(\mathbf{y}-(\mathbf{X}\otimes\mathbf{U}^{(2)}\otimes\mathbf{U}^{(1)})\boldsymbol{\beta})^{T}(\Sigma_{3}\otimes\Sigma_{2}\otimes\Sigma_{1})^{-1}(\mathbf{y}-(\mathbf{X}\otimes\mathbf{U}^{(2)}\otimes\mathbf{U}^{(1)})\boldsymbol{\beta})\\
 & =\arg\min_{\boldsymbol{\beta}}\boldsymbol{\beta}^{T}(\mathbf{X}^{T}\Sigma_{3}^{-1}\mathbf{X}\otimes\mathbf{U}^{(2)T}\Sigma_{2}^{-1}\mathbf{U}^{(2)}\otimes\mathbf{U}^{(1)T}\Sigma_{1}^{-1}\mathbf{U}^{(1)})\boldsymbol{\beta}-2\boldsymbol{\beta}^{T}(\mathbf{X}^{T}\Sigma_{3}^{-1}\otimes\mathbf{U}^{(2)T}\Sigma_{2}^{-1}\otimes\mathbf{U}^{(1)T}\Sigma_{1}^{-1})\mathbf{y}\\
 & =(\mathbf{X}^{T}\Sigma_{3}^{-1}\mathbf{X}\otimes\mathbf{U}^{(2)T}\Sigma_{2}^{-1}\mathbf{U}^{(2)}\otimes\mathbf{U}^{(1)T}\Sigma_{1}^{-1}\mathbf{U}^{(1)})^{-1}(\mathbf{X}^{T}\Sigma_{3}^{-1}\otimes\mathbf{U}^{(2)T}\Sigma_{2}^{-1}\otimes\mathbf{U}^{(1)T}\Sigma_{1}^{-1})\mathbf{y}\\
 & =(\mathbf{X}^{T}\Sigma_{3}^{-1}\mathbf{X})^{-1}\mathbf{X}^{T}\Sigma_{3}^{-1}\otimes(\mathbf{U}^{(2)T}\Sigma_{2}^{-1}\mathbf{U}^{(2)})^{-1}\mathbf{U}^{(2)T}\Sigma_{2}^{-1}\otimes(\mathbf{U}^{(1)T}\Sigma_{1}^{-1}\mathbf{U}^{(1)})^{-1}\mathbf{U}^{(1)T}\Sigma_{1}^{-1}\mathbf{y}
\end{align*}
Equivalently, this can be written in the tensor format as 
\[
\hat{\mathcal{B}}=\mathcal{Y}\times_{1}(\mathbf{U}^{(1)T}\Sigma_{1}^{-1}\mathbf{U}^{(1)})^{-1}\mathbf{U}^{(1)T}\Sigma_{1}^{-1}\times_{2}(\mathbf{U}^{(2)T}\Sigma_{2}^{-1}\mathbf{U}^{(2)})^{-1}\mathbf{U}^{(2)T}\Sigma_{2}^{-1}\times_{3}(\mathbf{X}^{T}\Sigma_{3}^{-1}\mathbf{X})^{-1}\mathbf{X}^{T}\Sigma_{3}^{-1}
\]
\end{proof}

\section{\label{sec:RegTenReg}The Proof of Proposition \ref{prop: twostepReg}}
\begin{proof}
$\boldsymbol{\beta}$ can be solved by

\begin{align*}
\hat{\boldsymbol{\beta}} & =\mathop{argmin}_{\boldsymbol{\beta}}(\mathbf{y}-(\mathbf{X}\otimes\mathbf{U}^{(2)}\otimes\mathbf{U}^{(1)})\boldsymbol{\beta})^{T}(\Sigma_{3}\otimes\Sigma_{2}\otimes\Sigma_{1})^{-1}(\mathbf{y}-(\mathbf{X}\otimes\mathbf{U}^{(2)}\otimes\mathbf{U}^{(1)})\boldsymbol{\beta})+P(\boldsymbol{\beta})\\
 & =\mathop{argmin}_{\boldsymbol{\beta}}\boldsymbol{\beta}^{T}(\mathbf{X}^{T}\Sigma_{3}^{-1}\mathbf{X}\otimes\mathbf{U}^{(2)T}\Sigma_{2}^{-1}\mathbf{U}^{(2)}\otimes\mathbf{U}^{(1)T}\Sigma_{1}^{-1}\mathbf{U}^{(1)})\boldsymbol{\beta}\\
 & +\boldsymbol{\beta}^{T}(\mathbf{X}^{T}\Sigma_{3}^{-1}\mathbf{X})\otimes(\lambda\mathbf{P}_{2}\otimes\mathbf{U}^{(1)^{T}}\mathbf{U}^{(1)}+\lambda\mathbf{U}^{(2)^{T}}\mathbf{U}^{(2)}\otimes\mathbf{P}_{1}+\lambda^{2}\mathbf{P}_{2}\otimes\mathbf{P}_{1})\boldsymbol{\beta}\\
 & -2\boldsymbol{\beta}^{T}(\mathbf{X}^{T}\Sigma_{3}^{-1}\otimes\mathbf{U}^{(2)T}\Sigma_{2}^{-1}\otimes\mathbf{U}^{(1)T}\Sigma_{1}^{-1})\mathbf{y}\\
 & =\mathop{argmin}_{\boldsymbol{\beta}}\boldsymbol{\beta}^{T}(\mathbf{X}^{T}\Sigma_{3}^{-1}\mathbf{X})\otimes(\mathbf{U}^{(2)^{T}}\Sigma_{2}^{-1}\mathbf{U}^{(2)}+\lambda\mathbf{P}_{2})\otimes(\mathbf{U}^{(1)^{T}}\Sigma_{1}^{-1}\mathbf{U}^{(1)}+\lambda\mathbf{P}_{1})\\
 & -2\boldsymbol{\beta}^{T}(\mathbf{X}^{T}\Sigma_{3}^{-1}\otimes\mathbf{U}^{(2)T}\Sigma_{2}^{-1}\otimes\mathbf{U}^{(1)T}\Sigma_{1}^{-1})\mathbf{y}\\
 & =((\mathbf{X}^{T}\Sigma_{3}^{-1}\mathbf{X}\otimes(\mathbf{U}^{(2)^{T}}\Sigma_{2}^{-1}\mathbf{U}^{(2)}+\lambda\mathbf{P}_{2})\otimes(\mathbf{U}^{(1)^{T}}\Sigma_{1}^{-1}\mathbf{U}^{(1)}+\lambda\mathbf{P}_{1}))^{-1}(\mathbf{X}^{T}\Sigma_{3}^{-1}\otimes\mathbf{U}^{(2)T}\Sigma_{2}^{-1}\otimes\mathbf{U}^{(1)T}\Sigma_{1}^{-1})\mathbf{y}\\
 & =(\mathbf{X}^{T}\Sigma_{3}^{-1}\mathbf{X})^{-1}\mathbf{X}^{T}\Sigma_{3}^{-1}\otimes(\mathbf{U}^{(2)^{T}}\Sigma_{2}^{-1}\mathbf{U}^{(2)}+\lambda\mathbf{P}_{2})^{-1}\mathbf{U}^{(2)^{T}}\Sigma_{2}^{-1}\otimes(\mathbf{U}^{(1)^{T}}\Sigma_{1}^{-1}\mathbf{U}^{(1)}+\lambda\mathbf{P}_{1})^{-1}\mathbf{U}^{(1)^{T}}\Sigma_{1}^{-1}\mathbf{y},
\end{align*}
which is equivalent to solve $\mathcal{B}$ in the tensor format as
shown in \eqref{eq: RegTensorRegsol}. 
\end{proof}

\section{The Proof of Proposition \ref{prop: onestepU}}

\label{sec: BCD} 
\begin{proof}
$\beta$ can be solved by
\begin{align*}
 & \mathop{\mathrm{argmin}}_{\boldsymbol{\beta}}(\mathbf{y}-(\mathbf{X}\otimes\mathbf{U}^{(2)}\otimes\mathbf{U}^{(1)})\boldsymbol{\beta})^{T}(\Sigma_{3}\otimes\Sigma_{2}\otimes\Sigma_{1})^{-1}(\mathbf{y}-(\mathbf{X}\otimes\mathbf{U}^{(2)}\otimes\mathbf{U}^{(1)})\boldsymbol{\beta})\\
= & \beta^{T}(\mathbf{X}\otimes\mathbf{U}^{(2)}\otimes\mathbf{U}^{(1)})^{T}(\Sigma_{3}\otimes\Sigma_{2}\otimes\Sigma_{1})^{-1}(\mathbf{X}\otimes\mathbf{U}^{(2)}\otimes\mathbf{U}^{(1)})\beta-2\beta^{T}(\mathbf{X}\otimes\mathbf{U}^{(2)}\otimes\mathbf{U}^{(1)})^{T}(\Sigma_{3}\otimes\Sigma_{2}\otimes\Sigma_{1})^{-1}y
\end{align*}
$\mathbf{y}$ is the vectorized $\mathcal{Y}$ with size $\mathbf{y}\in\mathbb{R^{\mathrm{n_{1}n_{2}N\times1}}}$,
$\boldsymbol{\mathbf{\beta}}$ is the vectorized $\mathcal{B}$ with
size $\boldsymbol{\mathbf{\beta}}\in\mathbb{R}^{pI_{1}I_{2}\times1}$.
If we optimize the $\beta$ gives
\begin{align*}
\hat{\beta} & =((\mathbf{X}^{T}\Sigma_{3}^{-1}\mathbf{X})^{-1}\mathbf{X}^{T}\Sigma_{3}^{-1}\otimes(\mathbf{U}^{(2)^{T}}\Sigma_{2}^{-1}\mathbf{U}^{(2)})^{-1}\mathbf{U}^{(2)^{T}}\Sigma_{2}^{-1}\otimes(\mathbf{U}^{(1)^{T}}\Sigma_{1}^{-1}\mathbf{U}^{(1)})^{-1}\mathbf{U}^{(1)^{T}}\Sigma_{1}^{-1})\mathbf{y}.\\
 & =((\mathbf{X}^{T}\Sigma_{3}^{-1}\mathbf{X})^{-1}\mathbf{X}^{T}\Sigma_{3}^{-1}\otimes\mathbf{U}^{(2)^{T}}\Sigma_{2}^{-1}\otimes\mathbf{U}^{(1)^{T}}\Sigma_{1}^{-1})\mathbf{y}
\end{align*}
Or equivalently 
\[
\hat{\mathcal{B}}=\mathcal{Y}\times_{1}\mathbf{U}^{(1)^{T}}\Sigma_{1}^{-1}\times_{2}\mathbf{U}^{(2)^{T}}\Sigma_{2}^{-1}\times_{3}(\mathbf{X}^{T}\Sigma_{3}^{-1}\mathbf{X})^{-1}\mathbf{X}^{T}\Sigma_{3}^{-1}.
\]
Plugging in the estimation of $\hat{\beta}$, we have 
\begin{align*}
 & (\mathbf{y}-(\mathbf{X}\otimes\mathbf{U}^{(2)}\otimes\mathbf{U}^{(1)})\boldsymbol{\hat{\beta}})^{T}(\Sigma_{3}\otimes\Sigma_{2}\otimes\Sigma_{1})^{-1}(\mathbf{y}-(\mathbf{X}\otimes\mathbf{U}^{(2)}\otimes\mathbf{U}^{(1)})\hat{\beta})\\
= & \hat{\beta}^{T}(\mathbf{X}\otimes\mathbf{U}^{(2)}\otimes\mathbf{U}^{(1)})^{T}(\Sigma_{3}\otimes\Sigma_{2}\otimes\Sigma_{1})^{-1}(\mathbf{X}\otimes\mathbf{U}^{(2)}\otimes\mathbf{U}^{(1)})\hat{\beta}-2\hat{\beta}^{T}(\mathbf{X}\otimes\mathbf{U}^{(2)}\otimes\mathbf{U}^{(1)})^{T}(\Sigma_{3}\otimes\Sigma_{2}\otimes\Sigma_{1})^{-1}y\\
= & \hat{\beta}^{T}((\mathbf{X}^{T}\Sigma_{3}^{-1}\mathbf{X})\otimes I\otimes I)\hat{\beta}-2\hat{\beta}^{T}(\mathbf{X}^{T}\Sigma_{3}^{-1}\otimes\mathbf{U}^{(2)^{T}}\Sigma_{2}^{-1}\otimes\mathbf{U}^{(1)^{T}}\Sigma_{1}^{-1})y\\
= & -\mathbf{y}^{T}(\Sigma_{3}^{-1}\mathbf{X}(\mathbf{X}^{T}\Sigma_{3}^{-1}\mathbf{X})^{-1}\mathbf{X}^{T}\Sigma_{3}^{-1})\otimes(\Sigma_{2}^{-1}\mathbf{U}^{(2)}\mathbf{U}^{(2)^{T}}\Sigma_{2}^{-1})\otimes(\Sigma_{1}^{-1}\mathbf{U}^{(1)}\mathbf{U}^{(1)^{T}}\Sigma_{1}^{-1})\mathbf{y}\\
= & -\|\mathbf{X}_{3}\otimes(\mathbf{U}^{(2)^{T}}\Sigma_{2}^{-1})\otimes(\mathbf{U}^{(1)^{T}}\Sigma_{1}^{-1})\mathbf{y}\|^{2}\\
= & -\|\mathcal{Y}\times_{1}\mathbf{U}^{(1)^{T}}\Sigma_{1}^{-1}\times_{2}\mathbf{U}^{(2)^{T}}\Sigma_{2}^{-1}\times_{3}\mathbf{X}_{3}\|^{2}
\end{align*}
\end{proof}
This close the first half of the proof. Then we will discuss how to
maximize $\|\mathcal{Y}\times_{1}\mathbf{U}^{(1)^{T}}\Sigma_{1}^{-1}\times_{2}\mathbf{U}^{(2)^{T}}\Sigma_{2}^{-1}\times_{3}\mathbf{X}_{3}\|^{2}$
according to the $\mathbf{U}^{(1)}$ and $\mathbf{U}^{(2)}$. We first
like to define that $\tilde{\mathbf{U}}^{(k)}=\Sigma_{k}^{-1/2}\mathbf{U}^{(k)}$.
The constraint is equivalent to $\tilde{\mathbf{U}}^{(k)T}\tilde{\mathbf{U}}^{(k)}=I$
and the data is$\mathbf{U}^{(k)^{T}}\Sigma_{k}^{-1}=\tilde{\mathbf{U}}^{(k)}\Sigma_{k}^{-1/2}$.
The problem becomes 
\begin{align*}
 & \|\mathcal{Y}\times_{1}\mathbf{U}^{(1)^{T}}\Sigma_{1}^{-1}\times_{2}\mathbf{U}^{(2)^{T}}\Sigma_{2}^{-1}\times_{3}\mathbf{X}_{3}\|^{2}\\
= & \|\mathcal{W}_{k}\times_{k}\tilde{\mathbf{U}}^{(k)}\Sigma_{k}^{-1/2}\|^{2}\\
= & \|\tilde{\mathbf{U}}^{(k)}\Sigma_{k}^{-1/2}\mathbf{W}_{k}\|^{2}
\end{align*}
Here $W_{k}$ is the $k^{th}$ mode unfolding of $\mathcal{W}_{k}=\mathcal{Y}\times_{i}\mathbf{U}^{(i)}\Sigma_{i}^{-1}\times_{3}X_{3}$.
It is not hard to prove that $\max_{\tilde{\mathbf{U}}^{(k)}}\|\tilde{\mathbf{U}}^{(k)}\Sigma_{k}^{-1/2}\mathbf{W}_{k}\|^{2}$
s.t. $\tilde{\mathbf{U}}^{(k)T}\tilde{\mathbf{U}}^{(k)}=I$ can be
solved by the first $P_{k}$ eigenvectors of $\Sigma_{k}^{-1/2}\mathbf{W}_{k}$.

\bibliographystyle{apalike}
\bibliography{cylinder}

\end{document}